\documentclass[10pt,twocolumn,letterpaper]{article}

\usepackage{cvpr}      %

\usepackage[normalem]{ulem}
\usepackage{subcaption}
\usepackage{caption,tabularx,booktabs}
\newcolumntype{?}{!{\vrule width 0.6pt}}
\usepackage{multicol}
\usepackage{multirow}
\usepackage{array}

\definecolor{cvprblue}{rgb}{0.21,0.49,0.74}
\usepackage[pagebackref,breaklinks,colorlinks,allcolors=cvprblue]{hyperref}

\title{Relightable Holoported Characters: Capturing and Relighting Dynamic Human Performance from Sparse Views}

\author{
Kunwar Maheep Singh$^{1}$ \and
Jianchun Chen$^{1,2}$ \and
Vladislav Golyanik$^{1}$ \and
Stephan J. Garbin$^{3}$ \and
Thabo Beeler$^{4}$ \and
Rishabh Dabral$^{1,2}$ \and
Marc Habermann$^{1,2}$ \and
Christian Theobalt$^{1,2}$
}

\begin{document}

\twocolumn[
\maketitle
\vspace{-30pt}
\begin{center}
\small
$^{1}$ Max Planck Institute for Informatics, Saarland Informatics Campus
\quad
$^{2}$ VIA Research Center
\quad
$^{3}$ Google, London
\quad
$^{4}$ Google, Zurich
\end{center}

]

\begin{abstract}
We present \emph{Relightable Holoported Characters} (RHC), a novel person-specific method for free-view rendering and relighting of full-body and highly dynamic humans solely observed from sparse-view RGB videos at inference.
In contrast to classical one-light-at-a-time (OLAT)-based human relighting, our transformer-based RelightNet predicts relit appearance within a single network pass, avoiding costly OLAT-basis capture and generation.
For training such a model, we introduce a new capture strategy and dataset recorded in a multi-view lightstage, where we alternate frames lit by random environment maps with uniformly lit tracking frames, simultaneously enabling accurate motion tracking and diverse illumination as well as dynamics coverage.
Inspired by the rendering equation, we derive physics-informed features that encode geometry, albedo, shading, and the virtual camera view from a coarse human mesh proxy and the input views.
Our RelightNet then takes these features as input and cross-attends them with a novel lighting condition, and regresses the relit appearance in the form of texel-aligned 3D Gaussian splats attached to the coarse mesh proxy.
Consequently, our RelightNet implicitly learns to efficiently compute the rendering equation for novel lighting conditions within a single feed-forward pass.
Experiments demonstrate our method’s superior visual fidelity and lighting reproduction compared to state-of-the-art approaches. See the \href{https://vcai.mpi-inf.mpg.de/projects/RHC/}{project page} for more details. %
\vspace{-8pt}

\end{abstract}
\vspace{-8pt}
\begin{figure}[t!]
    \includegraphics[width=\linewidth, trim=4.75cm 1.2cm 13cm 1.4cm, clip]{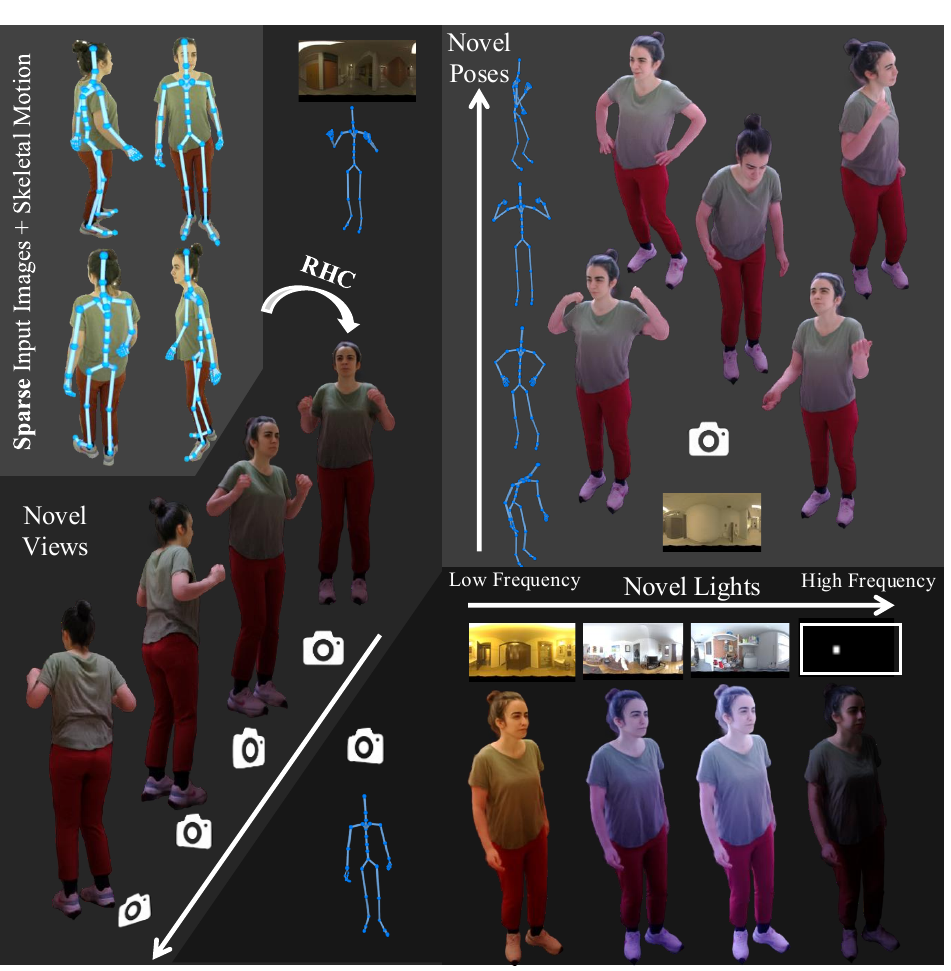}
    \vspace{-18pt}
    \caption{
    We present \emph{Relightable Holoported Characters}, the first method that takes sparse RGB images of a full-body human and generates photorealistic and relightable renderings, allowing for seamless placement of photoreal twins of real and dynamically moving humans into virtual environments. 
    }
    \vspace{-18pt}
    \label{fig:teaser}
\end{figure}
\vspace{-15pt}
\section{Introduction} \label{sec:introduction}
Reconstructing a human's geometry, appearance, as well as the scene lighting and rendering the person from a novel view and under novel lighting conditions---or \textit{human performance relighting} for short---is essential for seamless avatar insertion into virtual reality with applications in telepresence, film making, and gaming. 
However, accurately reconstructing physical illumination, or simulating it virtually, for a dynamically deforming human surface remains challenging due to complex light–matter interactions, which are difficult to model explicitly and computationally intractable to evaluate via the rendering equation~\cite{kajiya1986rendering}. 
To overcome these limitations, recent research tries to learn intrinsic material decomposition and relighting of the human appearance from image observations.
\par 
Some prior works~\cite{chen2022relighting4d,relightneuralactor2024eccv,chen2024meshavatar,wang2024intrinsicavatar,iqbal2023rana,xu2024relightable,xie2024ren,zhao2025surfel} rely on strong priors to intrinsically decompose and relight the avatar's appearance from monocular or multi-view video observed under \textit{static lighting}, often leading to implausible decompositions.
Lightstage-based methods~\cite{debevec2002lighting} enable better sampling of the lighting space, but exhaustively capturing every combination of light, view, and dynamically changing geometry is prohibitive. Consequently, many approaches focus on rigid parts (e.g., faces~\cite{bi2021deep,rao20253dpr,saito2024relightable,he2024diffrelight}, hands~\cite{iwase2023relightablehands,chen2024urhand,jiang20253d}) or replaying pre-captured performances~\cite{meka2020deep,debevec2000acquiring,he2024diffrelight,guo2019relightables}.
Methods that first capture an OLAT basis are similarly limited, as they are slow and only applicable for static scenes~\cite{debevec2000acquiring,rao20253dpr}. 
Relightable Full Body Gaussian Codec Avatars~\cite{wang2025relightable} generalizes relighting to full-body performances in unseen poses using randomly grouped OLAT lighting during training, but cannot leverage available image observations at inference to capture appearance and geometry consistent with the input images.
\par 
Motivated by the limitations of prior work, we introduce \textit{Relightable Holoported Characters (RHC)}, the first approach for photorealistic, controllable relighting of novel dynamic performances from sparse-view input images in a single feed-forward pass.
While trained on multi-view lightstage data, inference requires only four cameras capturing the subject under neutral illumination to \emph{relight unseen motions from novel virtual viewpoints}. 
Given sparse input images, we extract skeletal motion~\cite{thecaptury2020captury} and leverage a \textit{Character Animation Module}~\cite{habermann2021real} to deform a person-specific mesh template to coarsely match the subject’s geometry in the observed views.
Given the consistent UV parameterization of the deformed template mesh, appearance and relighting are then modeled in UV space using \emph{physics-informed features}, which our \textit{RelightNet} combines with an environment map to predict the relit appearance as texel-aligned 3D Gaussians.
These Gaussians are posed into global space and rasterized into 2D image space (Fig.~\ref{fig:main}).
Notably, instead of evaluating the rendering equation explicitly, RelightNet just requires a single feed-forward pass for human performance relighting at inference.
\par
Training RelightNet requires diverse lighting and motion coverage, along with frames that allow reliable geometry tracking. 
Thus, we introduce a new capture scheme and dataset recorded in a multi-view lightstage, tailored towards dynamic full-body relighting.
Traditional OLAT captures~\cite{debevec2000acquiring} require the subject to remain still and predict per-light outputs~\cite{he2024diffrelight}, which scales linearly with lighting conditions.
To efficiently sample diverse lighting, views, and dynamics, our new capture scheme interleaves environment maps simulating real-world illumination with uniformly lit tracking frames, enabling accurate geometry tracking while covering diverse lighting and dynamic motions.  
In summary, our contributions are as follows: 
\begin{itemize}
    \item \textbf{Relightable Holoported Characters:} the first method for photorealistic human performance relighting of unseen motions from sparse driving cameras at inference.
    \item \textbf{RelightNet and Physics-informed Feature Maps:} a transformer-based network, which efficiently computes the rendering equation in a single feed-forward pass, modeling light transport as cross-attention between environment maps and local physics-informed features encoding geometry, albedo, shading, and viewing direction.
    \item \textbf{Dataset and Capture Strategy:} A data capture strategy and a large-scale, multi-view lightstage dataset for dynamic full-body human relighting, featuring five subjects under more than 1000 natural lighting conditions, which we plan to publicly release as a benchmark. 
\end{itemize}
Our experiments demonstrate superior visual quality and lighting fidelity compared to state-of-the-art relightable full-body avatar approaches~\cite{chen2022relighting4d,wang2024intrinsicavatar,chen2024meshavatar}.
Our neural relighting model generalizes to unseen motions, views, and illuminations, including OLAT lighting conditions, despite not being trained on OLAT environments (Fig.~\ref{fig:teaser}). 
\vspace{-8pt}

\section{Related Work} \label{sec:related-work}
We do not review \emph{non-relightable} animatable avatars~\cite{junkawitsch2025eva, pang2024ash, zhu2024trihuman, guo2023vid2avatar, shen2023x, iandola2025squeezeme, bagautdinov2021driving, zielonka2025drivable, zhu2025ultra} or free-view human rendering~\cite{sun2025real, shetty2024holoported, jiang2025topology, jiang2025reperformer}, and instead focus on human relighting methods.
\par \noindent \textbf{Dynamic Human Relighting from Uncalibrated Lights.}
A vast majority of methods attempt to recover the relightable avatar from uncalibrated light sources~\cite{chen2022relighting4d,relightneuralactor2024eccv,chen2024meshavatar,wang2024intrinsicavatar,iqbal2023rana,xu2024relightable,xie2024ren,zhao2025surfel}, \textit{e.g.}, casual in-the-wild videos. 
From single or multi-view images, these approaches decompose environment lights with dynamic human geometry and material properties using inverse rendering~\cite{ramamoorthi2001signal}. 
However, the inherent ambiguities between geometry, material, and lighting are a long-standing challenge for general inverse rendering problems.
To make the task tractable, researchers utilized analytical material models, \textit{e.g.}, Microfact BRDF~\cite{relightneuralactor2024eccv,chen2022relighting4d}, Disney BRDF~\cite{wang2024intrinsicavatar}, or Lambertian BRDF~\cite{iqbal2023rana,xiao2024neca}.
Following the advances of neural radiance fields~\cite{mildenhall2021nerf}, the relightable avatar is typically represented by opacity, normal, and material parameters decoded from canonical space implicit functions, which are then deformed into pose space and rendered with physics-based rendering.
Some later works aim at improving the lighting and shadow correctness for unseen poses using hybrid representations of explicit meshes and implicit material fields~\cite{chen2024meshavatar}, a Hierarchical Distance Query algorithm~\cite{xu2024relightable}, or Monte Carlo ray tracing~\cite{wang2024intrinsicavatar}.
In sum, inverse rendering-based methods recover plausible geometry and material properties of full-body avatars from static, casual lighting.
However, the oversimplified BRDF model and their inability to compensate for tracking errors prevents them from achieving highly photorealistic relighting.
\par \noindent \textbf{Dynamic Human Relighting from Calibrated Lights.}
In the pursuit of photorealistic relightable human avatars, the Lightstage~\cite{debevec2000acquiring} was invented.
Lightstages enable precise illumination control via programmable LEDs, eliminating the aforementioned lighting ambiguity.  
By capturing OLAT data, one can composite and generate relit images of static objects in arbitrary lighting using the linearity of light transport.  
For dynamic human avatars, learning-based models primarily focus on individual body parts, \textit{e.g.}, hands~\cite{iwase2023relightablehands}, heads~\cite{bi2021deep}, and eyeglasses~\cite{li2023megane}, which are easier to track than full-body clothed humans.  
Later works improve rendering quality by incorporating 3D Gaussians~\cite{wang2025relightable}, but only consider sparse pose-conditioned input, failing to faithfully recreate dense wrinkle details observed in the input views.  
Diffusion models~\cite{he2024diffrelight} have also been used for face relighting, but complex non-rigid body deformations hinder extending these advances to full-body avatars.  
Volumetric capture systems~\cite{guo2019relightables,meka2020deep} achieve photorealistic relighting by combining high-resolution geometry with neural rendering or hybrid pipelines, but are limited to \emph{replaying} captured sequences and cannot generalize to novel poses from sparsely placed sensors.  
We instead target a generalizable relightable model that reconstructs full-body geometry and shading in unseen motions, leveraging high-frequency information from sparse input views at inference.
\par \noindent \textbf{Image-based Human Relighting.}
A prominent trend for relightable approaches is inspired by the success of generative models~\cite{ho2020denoising,rombach2022high,saharia2022photorealistic}.
Some researchers~\cite{zhang2025scaling,jin2024neural} aim for a general model across object categories and instances.
Relighting is conducted in 2D image space via image-to-image translation using generative models like Stable Diffusion~\cite{rombach2022high}.
When combined with lightstage data, they show promising performance in portrait relighting~\cite{pandey2021total,kim2024switchlight,mei2024holo}, shadow removal~\cite{yoon2024generative}, and harmonization~\cite{ren2024relightful} from single images.
These works motivate our end-to-end approach that fuses environment maps as tokens into image-space features.
However, prior diffusion-based relighting works~\cite{poirier2024diffusion} show image-space methods lack 3D and temporal consistency, limiting view-consistent tasks.
In contrast, we exploit explicit dynamic 3D geometry with efficient rasterization, producing temporally consistent relighting.
\vspace{-10pt}

\section{Dataset and Capture Strategy} \label{sec:dataset}
\begin{figure}[t!]
    \includegraphics[width=\linewidth]{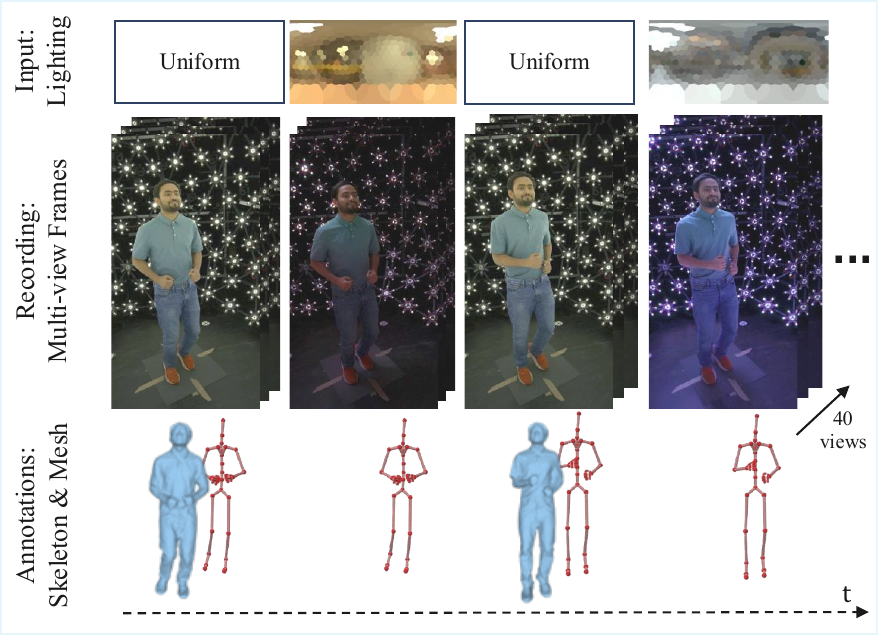}
    \vspace{-18pt}
     \caption{
     \textbf{Illustration of our data capture strategy.} 
     To learn a relightable full-body avatar, we propose to capture multi-view video sequences consisting of consecutive uniformly lit tracking frames and relit frames obtained by randomly projecting environment maps onto the lightstage LEDs.
    \vspace{-20pt}
     }
    \label{fig:data-capture}
\end{figure}

\begin{figure*}[ht]
\centering
    \includegraphics[width=1.0\textwidth, trim = 0.7cm 5.3cm 1.6cm 3.2cm, clip]{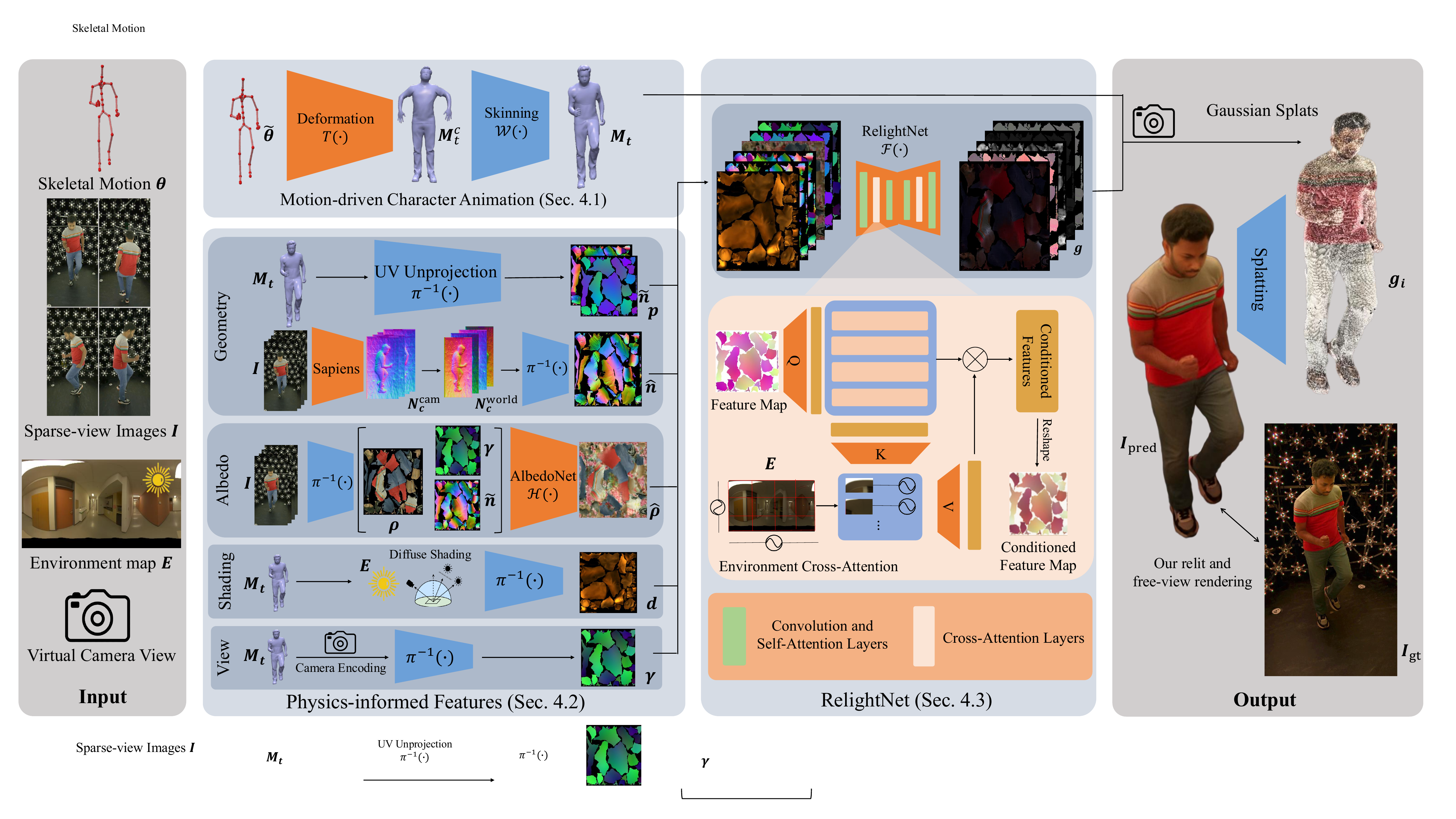}
    \vspace{-20pt}
    \caption{
    \textbf{Our approach, Relightable Holoported Characters (RHC).} Given four input views under flat lighting, skeleton pose, environment map, and camera parameters, our method generates photorealistic relighting.  
    First, a mesh-based avatar is animated using the skeleton pose (Sec.~\ref{sec:char_model}).  
    Physics-informed features (Sec.~\ref{sec:feat})—\textit{Geometry}, \textit{Albedo}, \textit{Shading}, and \textit{View Features}—are extracted from sparse-view images and mesh tracking, and fed into \textit{RelightNet} (Sec.~\ref{sec:relightnet}), which uses cross-attention to condition on the environment map.  
    \textit{RelightNet} predicts per-texel Gaussian parameters, which are placed on the mesh and splatted into the camera view.  
    }
    \vspace{-18pt}
    \label{fig:main}
\end{figure*}

Modeling a relightable full-body avatar from multi-view images requires precise separation of lighting, material, and geometry. 
To achieve this, we use a Lightstage system equipped with individually controllable LED panels and synchronized multi-view cameras, enabling active illumination control within the capture volume.
\par 
Existing data capture schemes for relightable faces~\cite{saito2024relightable} and hands~\cite{chen2024urhand} typically follow standard material acquisition practices. 
They rely on OLAT or grouped partially lit images interleaved with uniformly lit frames for geometry reconstruction and tracking.
While effective for localized regions such as faces and hands, extending this approach to dynamic full-body avatars adds further challenges. 
Synthesizing relit images of dynamic full-body avatars via linear composition requires multi-view OLAT images across diverse motions to sample the pose space, which is impractical to capture. 
Alternatively, capturing OLAT images during motion causes strong non-rigid clothing deformations.
As linear composition assumes static scenes, geometry errors accumulate and degrade relighting accuracy.
\par 
To address these challenges, we propose a novel data capture scheme for full-body relightable avatars. 
As illustrated in Fig.~\ref{fig:data-capture}, our setup alternates between two complementary frame types:
(1) frames illuminated with random environment maps, simulating diverse real-world lighting, 
and (2) uniformly lit frames for robust tracking. 
This interleaving ensures dense temporal alignment between geometry and illumination, enabling reliable motion tracking and relighting supervision, where the uniformly lit frames serve as approximate texture references for adjacent re-lit frames.
Compared to existing methods, our approach offers two main advantages: 
1) dense tracking frames that significantly improve robustness to non-rigid clothing deformations, and 
2) natural illumination captures that enable end-to-end training of our feedforward RelightNet without requiring analytical decomposition of material and lighting.

\vspace{-16pt}
\section{Method} \label{sec:method}
Our goal is to photorealistically render a digital twin of a real human performing a motion $\boldsymbol{\theta}$ -- observed only in sparse, calibrated multi-view images $\boldsymbol{I}$  -- from arbitrary viewpoints and lighting conditions $\mathbf{E}$ (see Fig.~\ref{fig:main}).
To learn our model, we assume lightstage data recorded according to our proposed strategy (Sec.~\ref{sec:dataset}). 
The core challenge lies in disentangling geometry, material, and illumination.
To address this, we introduce a mesh-based coarse character model that, given skeletal motion, predicts a non-rigidly deformed and posed mesh (Sec.~\ref{sec:char_model}).
In the consistent texture space of the tracked mesh, we encode physics-informed features relevant for evaluating the rendering equation (Sec.~\ref{sec:feat}).
Finally, these physics-informed features are fed into our \textit{RelightNet}, which cross-attends them with the environment map $\boldsymbol{E}$ to implicitly evaluate the rendering equation and produce relit appearance as Gaussian parameters $\boldsymbol{G}$ in a single feedforward pass (Sec.~\ref{sec:relightnet}).
\vspace{-8pt}
\subsection{Character Animation Module} \label{sec:char_model} 
We follow Deep Dynamic Characters (DDC)~\cite{habermann2021real} to pose a template mesh $\bar{\boldsymbol{M}}$ at timestep $t$, producing a motion-driven mesh $\boldsymbol{M}_t$.  
The canonical mesh
\vspace{-2pt}
\begin{equation}
    \boldsymbol{M}^c_t = T(\bar{\boldsymbol{M}}, \mathbf{a}_t, \mathbf{b}_t, \mathbf{\Delta}_t),
    \vspace{-4pt}
\end{equation}

is deformed in a coarse-to-fine manner, where $T(\cdot)$ is parameterized by embedded graph~\cite{sumner2007embedded} node rotations $\mathbf{a}_t$ and translations $\mathbf{b}_t$, together with fine vertex offsets $\mathbf{\Delta}_t$. 
Following DDC, we use two deformation networks
\vspace{-3pt}
\begin{equation}
    \mathbf{a}_t, \mathbf{b}_t = \mathcal{G}_\mathrm{eg}(\bar{\boldsymbol{M}}, \tilde{\boldsymbol{\theta}}_t), \quad
    \mathbf{\Delta}_t = \mathcal{G}_\mathrm{delta}(\bar{\boldsymbol{M}}, \tilde{\boldsymbol{\theta}}_t).
    \label{eq:ddc}
    \vspace{-3pt}
\end{equation}
The embedded-graph deformation network $\mathcal{G}_\mathrm{eg}$ predicts low-frequency node transformations, and the vertex refinement network $\mathcal{G}_\mathrm{delta}$ regresses high-frequency per-vertex displacements.
Here, $\tilde{\boldsymbol{\theta}}_t$ denotes the \emph{normalized motion} sequence $\{\boldsymbol{\theta}_i : i \in \{t\!-\!2,\ldots,t\}\}$. 
As in DDC, normalization removes global translation and global rotation from the tracked motion around the vertical axis.  
The posed mesh 
\vspace{-2pt}
\begin{equation}
    \boldsymbol{M}_t = \mathcal{W}(\boldsymbol{M}^c_t, \boldsymbol{\theta}_t, \boldsymbol{W}),
\end{equation}
is obtained by applying linear blend skinning (LBS), where $\boldsymbol{W}$ denotes the skinning weights.  
For training details, see the supplementary material.
\par 
This results in a temporally consistent mesh $\boldsymbol{M}_t$ with a stable UV parameterization, which offers two key benefits:
(1) it enables learning of relit appearance in canonical 2D texture space, simplifying training compared to learning it in 3D; 
and (2) it overcomes the challenge of tracking the template mesh in relit frames by leveraging the available skeleton tracking to animate the mesh.
\subsection{Physics-informed Features} \label{sec:feat}
\begin{figure*}[h!]
\centering
    \includegraphics[width=\textwidth, trim=15.25cm 13.8cm 6cm 5cm, clip]{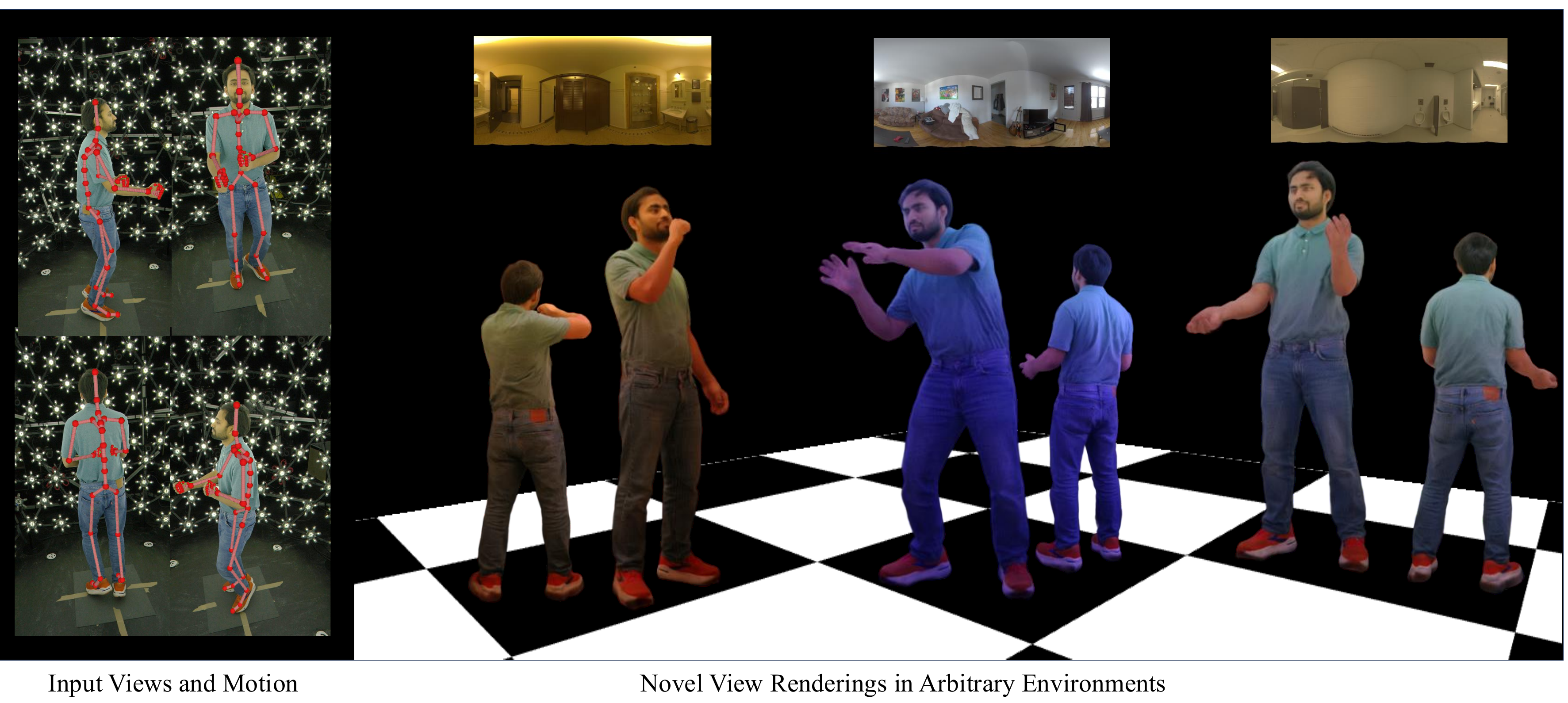}
    \vspace{-20pt}
    \caption{\textbf{Qualitative results.} Here, we show the 4 sparse input views of a person performing unseen motions. Our method, RHC, is then able to photorealistically render free-views under novel lighting conditions. Note that we also visualize different poses here, demonstrating the robustness to arbitrary skeletal motions.}
    \label{fig:qual}
    \vspace{-18pt}
\end{figure*}

Our goal now is to evaluate the rendering equation for a new lighting condition and for an unseen dynamic human performance in a single feed-forward pass using the consistent UV parameterization of the mesh $\boldsymbol{M}_t$.
The rendering equation~\cite{kajiya1986rendering} states that the outgoing radiance $\boldsymbol{L}_o(\boldsymbol{x},\boldsymbol{\omega}_o)$ is
\begin{equation}
\footnotesize{
\boldsymbol{\rho}(\boldsymbol{x}) \int_{\boldsymbol{\omega}_i} f_r(\boldsymbol{x},\boldsymbol{\omega}_i,\boldsymbol{\omega}_o)
\boldsymbol{L}_i(\boldsymbol{x},\boldsymbol{\omega}_i)
\boldsymbol{V}(\boldsymbol{x,\omega}_i)
\langle \boldsymbol{\omega}_i, \boldsymbol{n}\rangle  d\boldsymbol{\omega}_i,}
\label{eq:rendering_eq}
\end{equation}
where $\boldsymbol{x}$ is a surface point with normal $\boldsymbol{n}$. $\boldsymbol{\omega}_i$ and $\boldsymbol{\omega}_o$ denote the incoming and outgoing directions, $\boldsymbol{\rho}$ is the albedo, $f_r$ is the BRDF, $\boldsymbol{L}_i$ and $\boldsymbol{L}_o$ denote the incoming and outgoing light intensities, and $\boldsymbol{V}$ indicates visibility.
\par 
Directly recovering the albedo and BRDF for a deforming human mesh $\boldsymbol{M}$ is infeasible due to over-smoothed geometry and tracking noise, while simple BRDFs cannot model complex materials such as skin or fabric.
Our key idea is to encode an approximation of the individual components that define the rendering equation, i.e., geometry, albedo, shading, and view, in the 2D UV space of the mesh, which we call physics-informed features.
\par \noindent \textbf{Geometry Features.} %
Geometry features provide spatial cues essential for modeling shading and interreflections. 
Among them, surface normals capture local shape details that are critical for light interaction.
Our mesh normals provide a consistent coarse structure, and we use a 3-frame stack of mesh normals for temporal context
\vspace{-5pt}
\begin{equation}
\tilde{\boldsymbol{n}} = \{\boldsymbol{n}_{t-2}, \boldsymbol{n}_{t-1}, \boldsymbol{n}_t\}.
\end{equation}
However, they may miss fine details due to the comparably low mesh resolution and the limited capacity of the Character Animation Module.
Therefore, we also recover image normals $\boldsymbol{N}_c^\mathrm{cam}$ for all input views using Sapiens~\cite{khirodkar2025sapiens}.
We transform them to world space
\vspace{-5pt}
\begin{equation}
\boldsymbol{N}_c^\mathrm{world} = \boldsymbol{R}_c \boldsymbol{N}_c^\mathrm{cam},
\vspace{-2pt}
\end{equation}
and then unproject them to the mesh's UV space
\vspace{-5pt}
\begin{equation}
\hat{\boldsymbol{n}} = \sum_c \Pi^{-1}(\boldsymbol{N}_c^\mathrm{world}, \boldsymbol{M}),
\label{eq:unproj}
\vspace{-8pt}
\end{equation}
where $\Pi^{-1}$ denotes the unprojection operator. The resulting UV maps are averaged, following Holoported Characters (HPC)~\cite{shetty2024holoported}, to yield the final high-frequency normal feature $\hat{\boldsymbol{n}}$.
To capture near-field interreflections, we further encode the mesh position in a position map $\boldsymbol{p} = \Pi^{-1}(\boldsymbol{M}, \boldsymbol{M})$, storing global texel positions.
Together, $\tilde{\boldsymbol{n}}$, $\hat{\boldsymbol{n}}$, and $\boldsymbol{p}$ form the input to RelightNet responsible for encoding geometry.
\par \noindent \textbf{Albedo Features.} %
Albedo represents the intrinsic surface color, independent of lighting, and is essential to disentangle reflectance from illumination effects.
The appearance under uniform illumination approximates the surface albedo~\cite{mei2023lightpainter,mei2024holo,pandey2021total,mei2025lux}, which can be inferred from sparse-view RGB inputs.
We unproject the input images $\boldsymbol{I}$ into the template UV space to obtain an initial albedo map $\boldsymbol{\rho}$, similar to Eq.~\ref{eq:unproj}, which is able to compensate for tracking errors.
Due to limited views and self-occlusions, $\boldsymbol{\rho}$ contains only partial observations.
To refine it, we employ an albedo network, \emph{AlbedoNet}, defined as $\hat{\boldsymbol{\rho}} = \mathcal{H}(\boldsymbol{\rho}, \tilde{\boldsymbol{n}}, \boldsymbol{\gamma})$, which learns to inpaint missing information and to correct distorted regions.
Following HPC~\cite{shetty2024holoported}, the network takes a temporal normal stack $\tilde{\boldsymbol{n}}$ encoding geometry and the camera view $\boldsymbol{\gamma}$ encoded in texel space.
Precisely, the view map is computed as the per-texel direction from the position map $\boldsymbol{p}$ to the camera origin.
The refined albedo $\hat{\boldsymbol{\rho}}$ will later be the input to our RelightNet and is our physics-informed feature to approximate the true albedo in the rendering equation.
\par \noindent \textbf{Shading and View Features.} %
Physics-informed shading helps the network focus on high-frequency appearance rather than low-frequency illumination.
We compute pre-integrated diffuse shading $\boldsymbol{d}$ on the mesh $\boldsymbol{M}$ as
\vspace{-5pt}
\begin{equation}
\boldsymbol{d} = \int_{\boldsymbol{\omega}_i} \boldsymbol{L}_i(\boldsymbol{x},\boldsymbol{\omega}_i)
\boldsymbol{V}(\boldsymbol{x},\boldsymbol{\omega}_i)
\langle \boldsymbol{\omega}_i, \boldsymbol{n}\rangle \, d\boldsymbol{\omega}_i,
\label{eq:diffuse}
\vspace{-3pt}
\end{equation}
considering only direct environment lighting.
Our diffuse shading $\boldsymbol{d}$ and camera encoding $\boldsymbol{\gamma}$ form the last physics-informed features for RelightNet, responsible for approximating shading and view in the rendering equation.
\subsection{RelightNet} \label{sec:relightnet}

\begin{figure*}[ht]
    \centering
    \includegraphics[width=\textwidth, trim=0.1cm 3.5cm 1.0cm 2.4cm, clip]{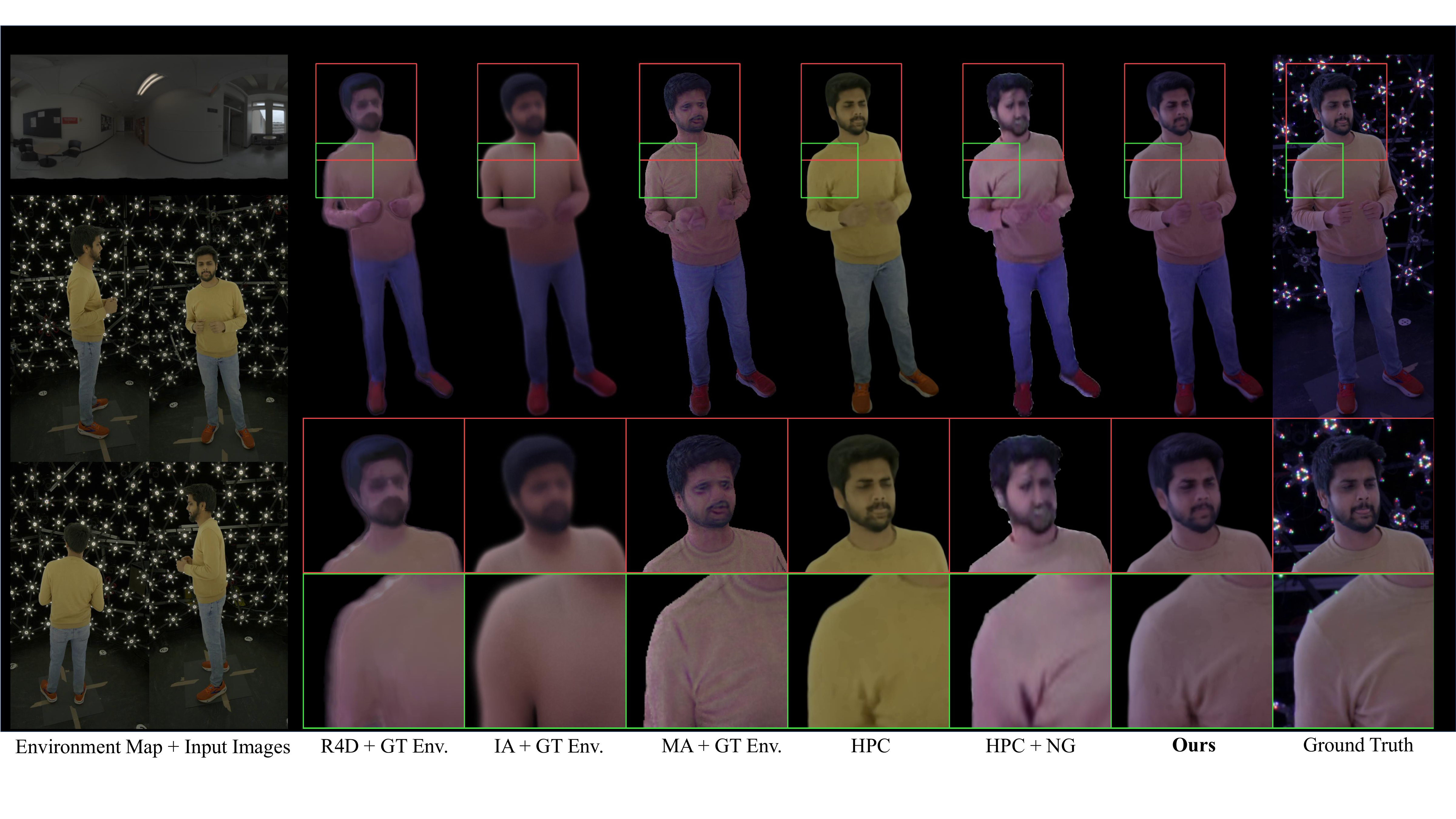}
    \vspace{-15pt}
    \caption{
    \textbf{Qualitative comparison.}
    We compare our method against state-of-the-art methods, including Relighting4D (R4D + GT Env.)~\cite{chen2022relighting4d} with ground-truth environment maps, as their original setting differs from our task.
    We also compare to advanced R4D variants, IA~\cite{wang2024intrinsicavatar} and MA~\cite{chen2024meshavatar}, augmented with ground-truth maps.
    Additionally, we compare to a sparse image-driven, non-relightable method, Holoported Characters (HPC)~\cite{shetty2024holoported}, and a relightable version (HPC + NG) using a recent image-based relighting network~\cite{jin2024neural}.
    Our method consistently outperforms all baselines across subjects and environment maps, highlighting its superior performance.
    }
    \label{fig:exp-main}
    \vspace{-15pt}
\end{figure*}

After defining the features relevant to evaluating the rendering equation, we now introduce RelightNet, an efficient 2D convolutional network, which learns the integration over lighting directions as well as the residual from approximate physics-informed features to real relit appearance from lightstage data captures.
In detail, given the physics-informed features in the UV space of our tracked mesh, we introduce \textit{RelightNet} $\mathcal{F}$
\vspace{-5pt}
\begin{equation}
    \boldsymbol{g} = \mathcal{F}(\boldsymbol{f}; \boldsymbol{E}) \quad \boldsymbol{f}=\{ \tilde{\boldsymbol{n}}, \hat{\boldsymbol{n}}, \boldsymbol{p}, \hat{\boldsymbol{\rho}}, \boldsymbol{d}, \boldsymbol{\gamma}\},
    \label{eq:relightnet}
    \vspace{-3pt}
\end{equation}
a network designed to translate a stack of normal, albedo, shading, and view features $\boldsymbol{f}$ into a realistic relit appearance texture $\boldsymbol{g}$ illuminated by an HDR environment map $\boldsymbol{E}$.
As shown in Fig.~\ref{fig:main}, \textit{RelightNet} combines convolution and self-attention layers in UV space with cross-attention layers that fuse the environment map into multi-scale local UV features.
The cross-attention formulation is inspired by the rendering equation (Eq.~\ref{eq:rendering_eq}), where each UV texel aggregates light contributions from all directions.
We linearly project the flattened environment map, concatenated with 2D sinusoidal positional encoding, into $(\boldsymbol{K}_\text{e}, \boldsymbol{V}_\text{e})$ and each texel feature into $\boldsymbol{Q}_\text{f}$, and perform multi-head cross-attention.
\par
Rather than directly predicting RGB texel colors, we predict parameters of relit 3D Gaussian splats~\cite{kerbl20233d}, following recent works on dynamic human modeling~\cite{Pang_2024_CVPR}.
Each UV texel corresponds to a 3D Gaussian with position $\boldsymbol{p}_i$, scaling $\boldsymbol{s}_i$, rotation $\boldsymbol{r}_i$, opacity $\boldsymbol{o}_i$, and color $\boldsymbol{c}_i$.
To facilitate learning, we initialize positions $\bar{\boldsymbol{p}}_i$ from the mesh surface $\boldsymbol{M}$ and scalings $\bar{\boldsymbol{s}}_i$ as mean distances to neighboring Gaussians in the first frame.
For each texel, we predict positional and scaling offsets $\delta\boldsymbol{p}_i$, $\delta\boldsymbol{s}_i$, along with $\boldsymbol{c}_i, \boldsymbol{r}_i, \boldsymbol{o}_i$, and obtain the final parameters as $\boldsymbol{p}_i = \bar{\boldsymbol{p}}_i + \delta\boldsymbol{p}_i$, $\boldsymbol{s}_i = \bar{\boldsymbol{s}}_i \odot \delta\boldsymbol{s}_i$, where $\odot$ denotes element-wise multiplication.
The predicted Gaussians are rendered using the camera parameters to obtain the final relit image $\boldsymbol{I}_\mathrm{pred}$.
Although diffuse shading and self-occlusion are already modeled in $\boldsymbol{d}$, the end-to-end \textit{RelightNet} learns to capture complete light transport—including specular reflections (via view encoding $\boldsymbol{\gamma}$ and cross-attention with $\boldsymbol{E}$) and subsurface scattering (via geometry encodings $\tilde{\boldsymbol{n}}, \boldsymbol{p}$ and self-attention in UV space). 
The Gaussian projection, rendering equations, and training details are provided in the supplementary material.

\vspace{-5pt}
\section{Experiments} \label{sec:experiment}
\vspace{-3pt}
\textbf{Dataset.}
We evaluate on five subjects wearing diverse clothes.
Each model component ($\mathcal{G}$, $\mathcal{H}$, $\mathcal{F}$) is trained per subject.
Of 40 captured cameras, 37 are used for training and 3 for testing.
Training uses 1,015 HDR environment maps from Laval Indoor~\cite{gardner2017learning}, and testing uses 8 novel ones.
Each subject has 28,420 training and 14,336 testing frames per camera, evenly split between uniformly lit and relit frames, captured at 60 Hz.
Test sequences were recorded separately and feature unseen motions.
\par \noindent \textbf{Metrics.}
\begin{table*}[!t]
    \centering

        \caption{
        \textbf{Quantitative evaluation.}
        We compare our method to prior methods for human performance relighting from uncalibrated lighting conditions (R4D~\cite{chen2022relighting4d}) as well as a variant (R4D + GT Env), where we provide the ground truth environment maps for training. 
        We further extend this training strategy to state-of-the-art methods IA~\cite{wang2024intrinsicavatar} and MA~\cite{chen2024meshavatar}.
        Moreover, we compare to non-relightable sparse free-view rendering methods (HPC~\cite{shetty2024holoported}) and a variant where we employ a recent foundational model~\cite{jin2024neural} for image-based relighting (HPC + NG).
        Note that we outperform all competing methods across subjects and metrics. %
    }
    \vspace{-5pt}
    \label{tab:main}
    
    \footnotesize 
    \setlength{\tabcolsep}{1.2pt} 
    \renewcommand{\arraystretch}{1.2} 
    \begin{tabular}{|c?c | c | c ?c |c |c?c|c|c?c|c|c?c|c|c |}
        \hline
        \multirow{2}{*}{Method} & \multicolumn{3}{c?}{Subject 1 (S1)} & \multicolumn{3}{c?}{Subject 2 (S2)} & \multicolumn{3}{c?}{Subject 3 (S3)} & \multicolumn{3}{c?}{Subject 4 (S4)} & \multicolumn{3}{c|}{Subject 5 (S5)} \\
        \cline{2-16}
         & PSNR $\uparrow$ & LPIPS $\downarrow$ & SSIM $\uparrow$ & PSNR $\uparrow$ & LPIPS $\downarrow$ & SSIM $\uparrow$ & PSNR $\uparrow$ & LPIPS $\downarrow$ & SSIM $\uparrow$ & PSNR $\uparrow$ & LPIPS $\downarrow$ & SSIM $\uparrow$ & PSNR $\uparrow$ & LPIPS $\downarrow$ & SSIM $\uparrow$ \\
         \hline
         R4D~\cite{chen2022relighting4d} & 26.09 & 14.24 & 84.89 & 28.68 & 10.94 & 85.01 & 26.04 & 14.48 & 86.43 & 28.24 & 13.68 & 86.03 & 27.14 & 12.36 & 82.83 \\
         R4D~\cite{chen2022relighting4d} + GT Env & 29.89 & 10.31 & 87.15 & 31.13 & 8.04 & \uline{87.08} & \uline{29.32} & 10.62 & \uline{87.67} & \uline{31.98} & 10.07 & \uline{87.90} & 29.11 & 8.48 & 84.04 \\
         \hline
         IA~\cite{wang2024intrinsicavatar} + GT Env & 27.25 & 18.25 & 81.39 & 28.87 & 15.43 & 82.07 & 26.14 & 22.52 & 79.07 & 29.50 & 18.46 & 82.91 & 28.61 & 16.94 & 80.35 \\
         \hline
         MA~\cite{chen2024meshavatar} + GT Env & 28.52  & 10.44  & 82.76  & 27.73  & 11.87  & 80.15  & 28.01  & 10.48  & 84.39  & 30.05  &  \uline{9.55} & 84.14  & 30.01 & 8.42 & 84.20 \\
         \hline
         HPC~\cite{shetty2024holoported} & 25.84 & 11.41 & \uline{88.04} & 27.34 & 9.58 & 85.62 & 24.46 & 12.47 & 86.93 & 27.34 & 10.27 & 86.24 & 27.08 & 9.00 & \uline{84.37} \\
         HPC~\cite{shetty2024holoported} + NG~\cite{jin2024neural} & \uline{30.52} & \uline{8.75} & 87.49 & \uline{31.90} & \uline{7.99} & 82.62 & \textbf{29.80} & \uline{9.06} &  82.02 & 31.63 & 10.66 & 81.69 & \uline{30.52} & \uline{8.41} & 80.41 \\
         \hline
        \textbf{Ours} & \textbf{31.38} & \textbf{7.01} & \textbf{90.00} & \textbf{32.48} & \textbf{5.69} & \textbf{90.36} & 29.09 & \textbf{8.13} & \textbf{88.71} & \textbf{32.44} & \textbf{7.11} & \textbf{89.40} & \textbf{32.07} & \textbf{5.55} & \textbf{89.34} \\
        \hline 
    \end{tabular}
    \vspace{-10pt}

\end{table*}

\begin{table}[t]
    \footnotesize 
    \centering
    \caption{
    \textbf{Ablation study.}
    We ablate different design choices of our method.
    Removing individual components degrades metrics, confirming their importance. Skeleton tracking from sparse views shows minimal decrease, mainly due to hand tracking errors. Fewer input views also degrade results because of increased occlusion. Finally, OLAT data capture accumulates errors from each OLAT rendering.
    \vspace{-5pt}
    }
    \label{tab:ablation}
    
    \setlength{\tabcolsep}{1.2pt} 
    \renewcommand{\arraystretch}{1.2} 
    \begin{tabular}{|l?c|c|c|}
         \hline
         \hspace{2pt}Method & PSNR $\uparrow$ & LPIPS  $\downarrow$ &SSIM $\uparrow$\\
         \hline
         \hspace{2pt}Ours w/o geometry features & 31.73 & 5.58  & 89.01 \\
         \hspace{2pt}Ours w/o coarse normals & 31.96 & \textbf{5.55} & 89.12 \\
         \hspace{2pt}Ours w/o high-freq. normals feature & 31.94 & \textbf{5.55} & 89.18 \\
         \hspace{2pt}Ours w/o position map & 31.70 & 5.58  & 89.18  \\   
         \hline
         \hspace{2pt}Ours w/o albedo feature & 31.82 & 5.78  & 88.53 \\
         \hspace{2pt}Ours w/o diffuse shading feature & 31.59 & 5.74 & 88.80 \\  
         \hspace{2pt}Ours w/o camera encoding & 31.52 & 5.68 & 88.77 \\
         \hline
         \hspace{2pt}Ours w/o cross attention & 31.88  & \uline{5.56} & 89.18  \\
         \hline
         \hspace{2pt}Ours w/ sparse view skeleton tracking & 30.72 & 6.06 & 87.83 \\         
         \hline
         \hspace{2pt}Ours w/ 0 input views (pose only) & 31.39 & 6.23 & 87.50 \\
         \hspace{2pt}Ours w/ 2 input views & \uline{32.00} & 5.74 & 89.20 \\
         \hline
         \hspace{2pt}OLAT data capture & 26.42 & 10.46  & \textbf{89.86} \\
         \hline
         \hspace{2pt}\textbf{Ours} & \textbf{32.07} & \textbf{5.55} & \uline{89.34} \\
         \hline

    \end{tabular}
    \vspace{-15pt}

\end{table}

To evaluate our method, we report PSNR, SSIM~\cite{wang2004image}, and LPIPS~\cite{zhang2018perceptual} averaged over every 10th frame across all three test views.
\par \noindent \textbf{Baseline.}
Since no existing method directly addresses sparse-view relightable avatars, we adapt two types of baselines.  
First, we extend Relighting4D (R4D)~\cite{chen2022relighting4d} to a motion-driven, relightable avatar. The original R4D is trained on uniform-light tracking frames; we train it on relit frames with ground-truth environment maps for fair comparison.  
We also compare to IntrinsicAvatar (IA)~\cite{wang2024intrinsicavatar} and MeshAvatar (MA)~\cite{chen2024meshavatar}, trained similarly with ground-truth environment maps.  
We do not compare to Relightable Full Body Gaussian Codec Avatars~\cite{wang2025relightable} since their code is unavailable.
Second, we adapt HPC~\cite{shetty2024holoported}, a sparse-view photorealistic avatar method, by applying a per-identity finetuned Neural Gaffer (NG)~\cite{jin2024neural} for relighting.  

To ensure a fair relighting comparison, all methods are driven by the same high-quality skeletal motion extracted from our multi-view setup, thereby attributing performance differences only to the relighting models.
However, our ablation (Sec.~\ref{sec:exp_ablation}) demonstrates that even sparse-view motion capture creates visually plausible results.
All methods use the same dense views as our method for training.  
\par \noindent \textbf{Qualitative Results.}
In Fig.~\ref{fig:qual}, we show the ability of our method to generate photorealistic renderings in different illuminations, from novel viewpoints, and under novel poses. 
\subsection{Comparisons on Novel Sequences and Lighting} \label{sec:exp_main}
\textbf{Quantitative and Qualitative Comparisons.} Tab.~\ref{tab:main} shows that our method outperforms all baselines across subjects.  
Original R4D, trained on uniform-lit frames, slightly surpasses the non-relightable HPC in numerical metrics. When trained on relit frames with ground-truth environment maps (R4D + GT Env.), performance improves substantially, highlighting the effectiveness of our lightstage capture scheme.  
IA + GT Env. achieves similar quality to R4D + GT Env., limited by SMPL-based tracking, 3D representation, and BRDF assumptions. MA + GT Env. achieves sharper results than IA + GT Env. due to its hybrid explicit mesh and implicit material field representation but still suffers from artifacts from pose-only input and BRDF assumptions. HPC preserves sharp visual details from sparse-view input, but performance remains limited by the non-relightable representation; per-identity finetuned NG improves shading but still falls short of photorealism.  

Fig.~\ref{fig:exp-main} illustrates qualitative comparisons. Our method recovers sharp wrinkles and realistic shading compared to the ground truth. R4D + GT Env. produces blurrier results due to coarse geometry. IA/MA + GT Env. improve shading, but cannot capture realistic skin tone or cloth scattering. HPC synthesizes novel views with sharp details under uniform lighting, but NG relighting introduces inconsistent shading and disrupts fine structures (e.g., on the face).  
Additional results are shown in the supplementary material.
\vspace{-2pt}
\subsection{Ablation Studies} \label{sec:exp_ablation}
\begin{figure}[t!]
    \includegraphics[width=\linewidth, trim=0.0cm 4.7cm 0.0cm 0.0cm, clip]{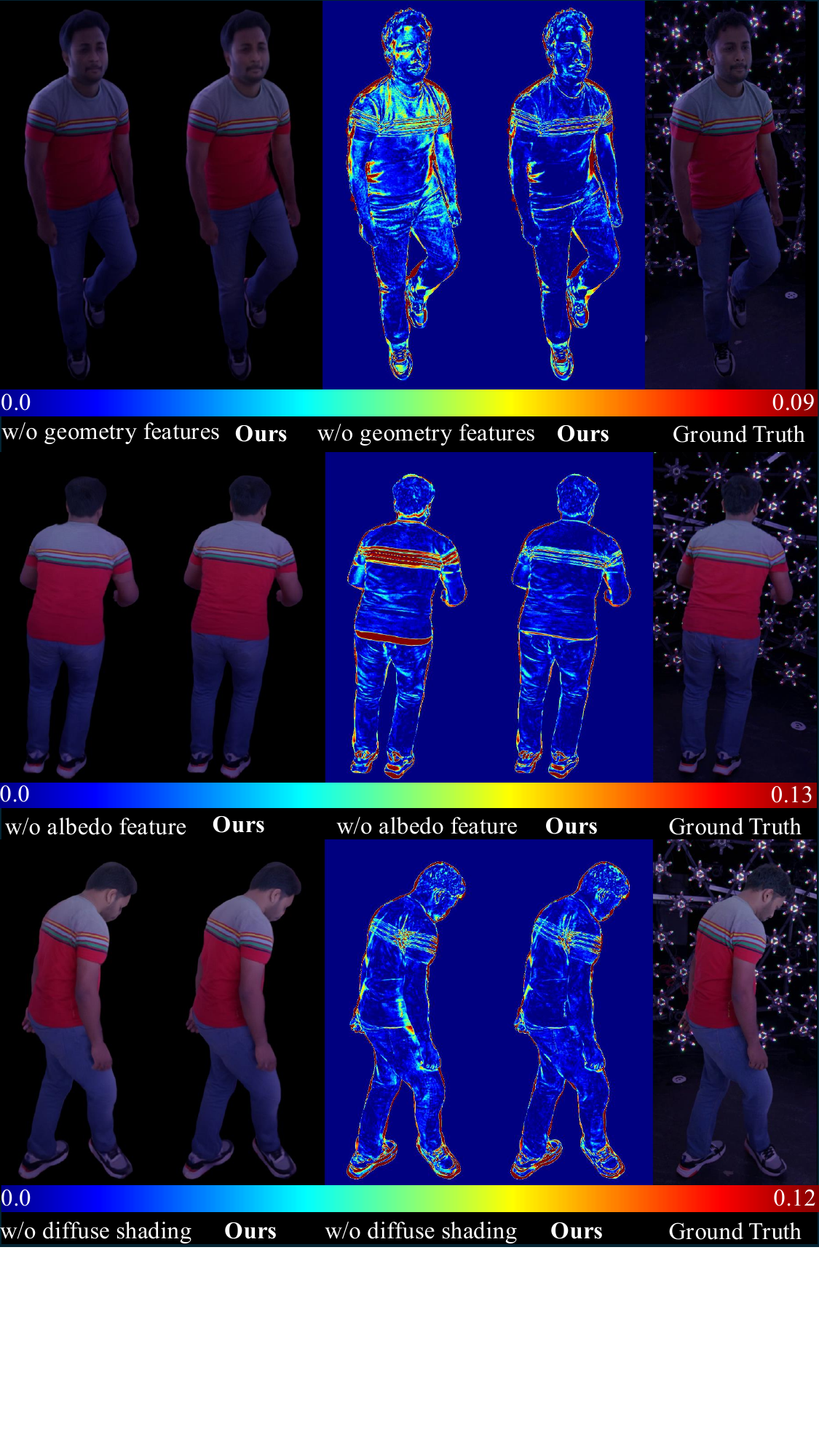}
    \vspace{-18pt}
    \caption{\textbf{Ablation of key design components.} Removing our geometry features hinders the learning of pose-dependent effects and also leads to reduced wrinkle fidelity due to missing high-frequency geometry. Excluding the albedo feature causes texture drift from tracking errors. Without diffuse shading, the model fails to capture self-shadows correctly.
    }

     \vspace{-15pt}
    \label{fig:exp-ablation-1}
\end{figure}

We further conduct quantitative (Tab.~\ref{tab:ablation}) and qualitative (Fig.~\ref{fig:exp-ablation-1}) ablation studies to validate our design choices.
\par \noindent \textbf{Geometry Features.}
Removing geometry features, $\boldsymbol{\hat{n}}$, $\tilde{\boldsymbol{n}}$, and $\boldsymbol{p}$,  deteriorates rendering quality significantly since we remove the model's ability to correct for pose-dependent errors (Fig.~\ref{fig:exp-ablation-1}). This is also reflected in the metrics in Tab.~\ref{tab:ablation}.
\par \noindent \textbf{Coarse Normals.}
Without the coarse normal stack, the network lacks temporal context and misses temporal effects.
\par \noindent \textbf{High-frequency Normal Features.}
The additional high-frequency normals $\boldsymbol{\hat{n}}$ improve the rendering quality of wrinkles and fine details, leading to better metrics. 
\par  \noindent \textbf{Position Map.}
The position map reduces the model's burden of having to learn the surface curvature. Removing it also removes the ability to model correct light transport from finitely far away light sources.
\par  \noindent \textbf{Albedo Features.}
The removal of albedo features $\hat{\boldsymbol{\rho}}$ degrades the metrics since it fails to correct texture drift due to incorrect tracking, leading to errors in rendering (Fig.~\ref{fig:exp-ablation-1}).
\par  \noindent \textbf{Diffuse Shading.}
Removing the diffuse shading feature eliminates explicitly modeled self-shadow, deteriorating rendering quality in occluded regions. The ablated baseline fails to produce correct self-shadows (Fig.~\ref{fig:exp-ablation-1}).
\par  \noindent \textbf{Camera Encoding.}
Without camera encoding $\boldsymbol{\gamma}$, our method cannot model view-dependent effects.
\par  \noindent \textbf{Cross Attention.}
We evaluate the effectiveness of cross-attention for conditioning the environment map.
The baseline without cross-attention concatenates the resized environment map $\boldsymbol{E}$ with UNet input features.
\par  \noindent \textbf{Dense View Skeleton Tracking.}
In the actual use case, the input motion is captured from only 4 views.
Here, we study our model's performance with skeleton tracking from sparse views and find a slight performance drop, mostly due to imperfect hand tracking, while the visual quality is preserved.

\par  \noindent \textbf{Number of input views.}
A reduction in the number of input views results in a decline in detail preservation as more details are occluded from our model. This leads to a performance drop as shown in Tab.~\ref{tab:ablation}. Note, we still outperform all baselines with just our pose-only model.

\par
\begin{figure}[t!]
    \includegraphics[width=\linewidth, trim=19cm 2cm 24cm 27cm, clip]{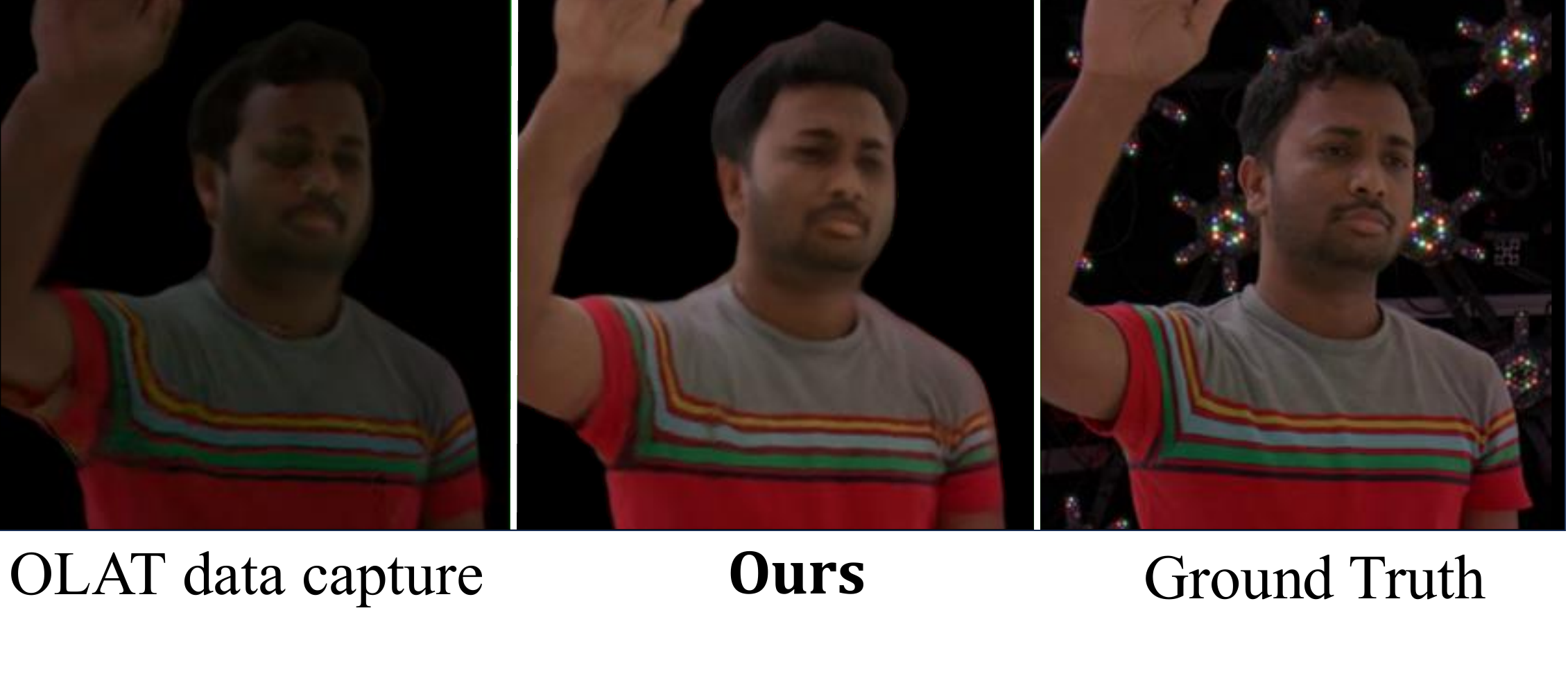}
    \vspace{-28pt}
     \caption{
     \textbf{Effect of different capture strategies.}
     Learning OLATs in an end-to-end manner and linearly combining them to illuminate to natural environments is slow and accumulates errors.
    \vspace{-15pt}
     }
    \label{fig:olat}
\end{figure}

 \noindent \textbf{Training with OLAT Capture.}
To validate our capture strategy, we record a dynamic grouped OLAT sequence for one subject and train an end-to-end relighting model to predict grouped OLATs, which are linearly combined to simulate environment lighting. Details are provided in the supplementary material.
Visual quality degrades due to accumulated per-light prediction errors (Fig.~\ref{fig:olat}).

\vspace{-5pt}

\vspace{-2pt}
\section{Limitations and Future Work}\label{sec:limitations} 

While RHC provides the first solution for photorealistic relighting of unseen human performances from sparse views, it has several limitations suggesting directions for future work.
Our method is trained per identity and does not generalize across subjects or clothing; extending it with priors from large generative models~\cite{rombach2022high,ho2020denoising} could improve scalability.
RHC also struggles with topology changes (e.g., removing a jacket), translucent materials, and accessories such as glasses, while some specular highlights on body parts (eyes, fingernails) are missed.
These challenges could be mitigated by combining RHC with a layered clothing representation~\cite{rong2025gaussian} and more diverse training data to improve robustness across material types.
Finally, the current pipeline prioritizes visual quality over runtime efficiency ($\approx$2~FPS); integrating CUDA-accelerated ray tracing~\cite{parker2010optix} and lightweight neural components via distillation~\cite{hinton2015distilling} could enable interactive high-quality relighting.

\vspace{-4pt}
\section{Conclusion}\label{sec:conclusion} 

Relightable Holoported Characters (RHC) marks a step change in human relighting—from static, replay-based, or pose-driven avatars that hallucinate appearance details, to dynamic, photorealistic humans that reconstruct real, pose-dependent wrinkles and shading from sparse camera views. 
By showing that end-to-end learning from natural lightstage captures can reproduce fine, view-consistent detail and shading, without OLAT data, RHC makes scalable, relightable telepresence feasible. 
This work paves the way for identity-generalizable, scene-adaptive avatars that move seamlessly between the real and virtual worlds.

\section{Acknowledgements}
This work was funded by the Saarbr\"ucken Research Center for Visual Computing and Artificial Intelligence (VIA).

{
    \small
    \bibliographystyle{ieeenat_fullname}
    \bibliography{main}
}
\clearpage

\appendix

\section{Overview}
In this supplemental document, we provide further details on the pre-scanned character model (Sec.~\ref{sec:s-ddc}).  
Next, we describe how the diffuse features are computed (Sec.~\ref{sec:s-diffuse}).  
We then present a more in-depth explanation of our cross-attention mechanism (Sec.~\ref{sec:s-cross-att}), Gaussian Projection (Sec.~\ref{sec:s-gauss}), and network architectures (Sec.~\ref{sec:s-relightnet}).  
Following this, implementation details, training procedures, and loss functions are discussed (Sec.~\ref{sec:s-loss}).  
Finally, we detail the data capture and hardware setup (Sec.~\ref{sec:s-data}) and the baselines (Sec.~\ref{sec:s-baseline}), provide additional information on our dynamic OLAT experiment (Sec.~\ref{sec:s-olat}), give a runtime and memory breakdown along with comparisons to baselines (Sec.~\ref{sec:s-runtime}), and present further ablations (Sec.~\ref{sec:s-fur_abl}) as well as additional qualitative results (Sec.~\ref{sec:s-qual}).
\section{Character Animation Model}\label{sec:s-ddc}
For each subject defined in Sec. 4.1 in the main paper, we estimate the skeletal motion using dense camera views with Captury \cite{thecaptury2020captury}.
In addition, we use a body scanner to obtain a template mesh per subject with $4890$ vertices and $9700$ faces, which is later simplified as an embedded graph with $489$ nodes.
Paired with the template mesh, a person-specific skeleton is produced with $173$ joints, driven by the motion parameters $\boldsymbol{\theta}$ with $107$ degrees of freedom.
We generate the skinning weight $\boldsymbol{W}\in \mathbb{R}^{4890\times68}$ from Blender~\cite {blender} autorig based on a subset of body and hand joints.
\par
We train the Character Animation Networks $\mathcal{G}_\mathrm{eg}$ and $\mathcal{G}_\mathrm{delta}$ %
with multi-view images and NeuS2 \cite{wang2023neus2} supervisions in the tracking frames.
They produce 3D geometry interpolation in relit frames for training the neural relightable model, as well as the extrapolated geometry for novel motions during inference. %
Moreover, we annotate 20 part segmentation labels on the mesh vertices using ~\cite{li2020self} and assign rigidity weights, which facilitates the Laplacian and ARAP loss in training the \textit{Character Animation Network}.

\section{Diffuse Shading}\label{sec:s-diffuse}
\begin{figure}[t!]
    \includegraphics[width=\linewidth]{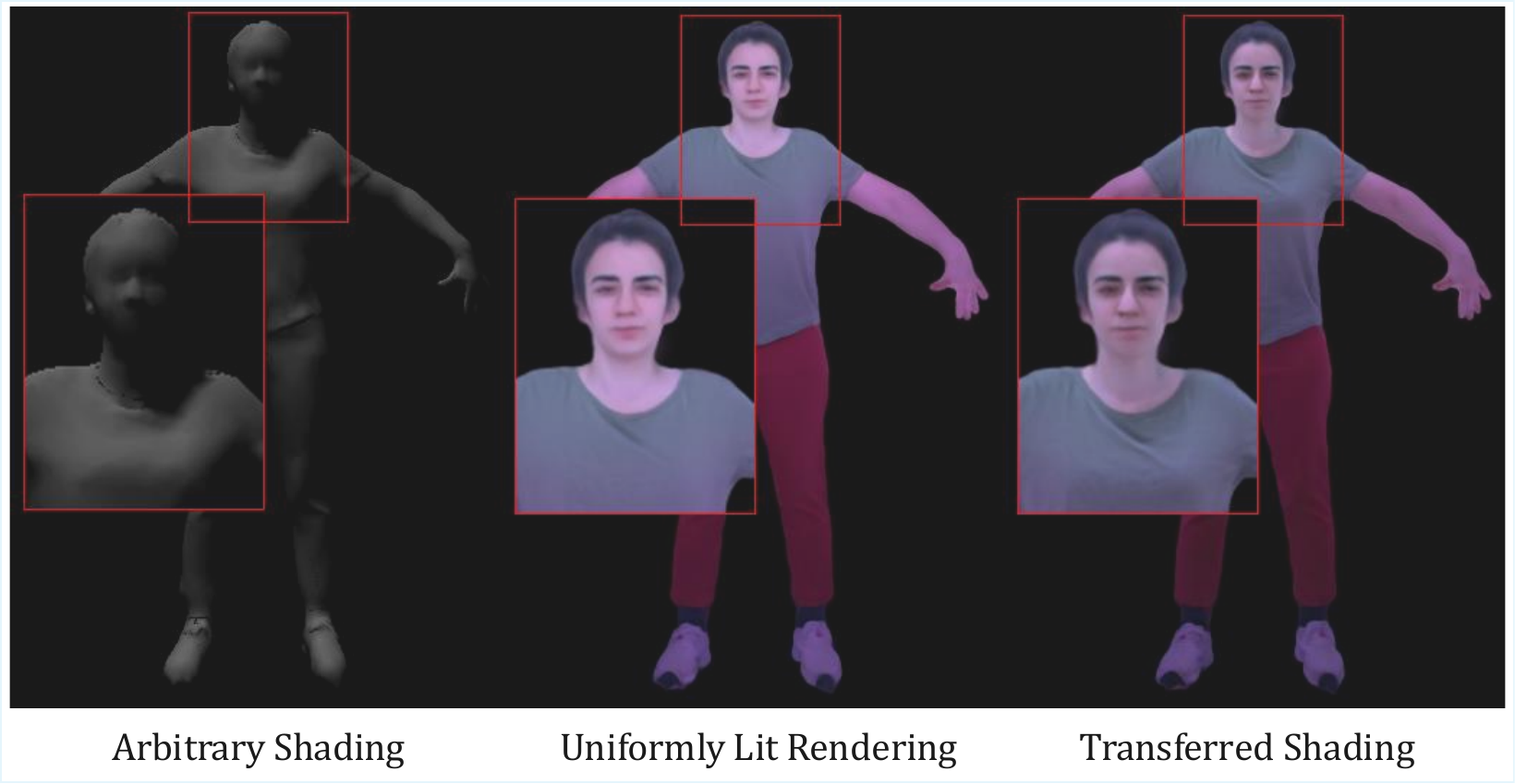}
    \vspace{-18pt}
     \caption{
     Our model enables the transfer of arbitrary shading to the rendered output. \textbf{Left}. diffuse shading computed under an arbitrary light source. \textbf{Middle}. uniformly lit rendering predicted by our model. \textbf{Right}. rendering obtained by providing the computed shading instead of the uniform diffuse shading. We note the correct transfer of shadows from our shading feature to the model rendering, showcasing the disentangled control our model provides.
    \vspace{-12pt}
     }
    \label{fig:diffuse_abl}
\end{figure}
The diffuse shading feature $\boldsymbol{d}$ is computed via a CPU-based one-bounce ray tracer from Open3D~\cite{zhou2018open3d}.
We first map the light sources from the environment map to the light emitters in the lightstage with pre-calibrated 3D positions of LEDs.
Then, we use this position to calculate the direction of incoming light for diffuse shading and occlusions. 
Jointly with the 3D position map $\boldsymbol{p}$, the shading features encode signals about the light-to-body distance, which particularly helps to recreate near-field lighting effects.

We show the disentangled control this shading allows in the final rendering in Fig.~\ref{fig:diffuse_abl}. Here, we use the diffuse shading from an OLAT light condition to replace the one in a uniformly lit rendering, and notice the shadowing effects in the final rendering exactly mirror those from the diffuse shading.
\section{Cross-attention Mechanism}\label{sec:s-cross-att}
The cross-attention layer processes the feature descriptor $\boldsymbol{h}_i$ corresponding to texel $i$ in the higher-layer feature map of \textit{RelightNet} with the RGB intensity value $\boldsymbol{e}_j$ of pixel $j$ of the environment map $\boldsymbol{E}$.
Let $\boldsymbol{q}$ denote the 2D coordinate of pixel $j$.
The intensity value $\boldsymbol{e}_j$ is first concatenated with 64 dimensional positional encoding $P(\boldsymbol{q})$~\cite{vaswani2017attention}, then linearly mapped into $\boldsymbol{K}_{\boldsymbol{e}}$, $\boldsymbol{V}_{\boldsymbol{e}}$.
\begin{equation}
    \boldsymbol{K}_{\boldsymbol{e}} = \boldsymbol{W}_K[\boldsymbol{e}_j,P(\boldsymbol{q})], \quad \boldsymbol{V}_{\boldsymbol{e}} = \boldsymbol{W}_V[\boldsymbol{e}_j,P(\boldsymbol{q})]
\end{equation}
Similarly, the feature descriptor $\boldsymbol{h}_i$ is linearly transformed into query vector $\boldsymbol{Q}_{\boldsymbol{f}}$:
\begin{equation}
    \boldsymbol{Q}_{\boldsymbol{f}} = \boldsymbol{W}_Q \boldsymbol{h}
\end{equation}
where $\boldsymbol{\boldsymbol{W}_{\{Q,K,V\}}}$ are the learnable weights.
The integration of positional encoding with environment lights ensures that the spatial relationships within the environment map are preserved during the attention process.

\section{Gaussian Projection and Rendering}\label{sec:s-gauss}
Each 3D Gaussian, predicted by our RelightNet, is defined by its center $\boldsymbol{p}_i$, scaling $\boldsymbol{s}_i$, rotation $\boldsymbol{r}_i$, opacity $\boldsymbol{o}_i$, and color $\boldsymbol{c}_i$.
Following EWA splatting~\cite{zwicker2002ewa}, we project each Gaussian into the image plane, obtaining a 2D center $\boldsymbol{\mu}_i$ and covariance matrix $\boldsymbol{\Sigma}_i$ as
\begin{equation}
\boldsymbol{\Sigma}_i = \boldsymbol{J}_i \boldsymbol{W}_i \boldsymbol{R}_i \mathrm{diag}(\boldsymbol{s}_i)\mathrm{diag}(\boldsymbol{s}_i)^{\!\top}\boldsymbol{R}_i^{\!\top}\boldsymbol{W}_i^{\!\top}\boldsymbol{J}_i^{\!\top},
\end{equation}
where $\boldsymbol{J}_i$ is the Jacobian of the viewing transformation, $\boldsymbol{W}_i$ the camera-to-world matrix, and $\boldsymbol{R}_i$ the rotation from quaternion $\boldsymbol{r}_i$.
Given the 2D Gaussian $G(\boldsymbol{x}; \boldsymbol{\mu}_i, \boldsymbol{\Sigma}_i)$, the color of pixel $p$ is computed as a weighted alpha-composited sum:
\begin{equation}
    \mathbf{C}_p = \sum_{j \in \mathcal{N}} \mathbf{c}_j \alpha_j \prod_{k=1}^{j-1} (1 - \alpha_k),
    \label{eq:gaussian_render_supp}
\end{equation}
where $\alpha_j$ denotes the transparency of the $j$-th Gaussian, modulated by its contribution $G(\boldsymbol{x}_p; \boldsymbol{\mu}_j, \boldsymbol{\Sigma}_j)$.
This formulation ensures smooth, view-consistent appearance reconstruction across multiple viewpoints.

\section{RelightNet}\label{sec:s-relightnet}
RelightNet consists of interleaved convolutional, self-attention, and cross-attention layers for its encoder and decoder. 
Due to the high computational cost of self-attention and cross-attention at high resolution ($\ge 64\times64$), we only use convolutional layers at these resolutions. 
For the higher layers of the \textit{RelightNet} with low-resolution feature maps, we follow every convolutional layer with a self-attention layer and a cross-attention layer between the UV space features and the environment map. 
Here, self-attention is computed between flattened elements of the feature map.
The number of heads in self-attention and cross-attention is set as $4$.
Table~\ref{tab:s-na} illustrates the concrete architecture of \textit{RelightNet}. 
\begin{table}[h!]
    \centering
    \caption{\textbf{Illustration of the \textit{RelightNet} architecture.} In the operation column, "C" denotes a convolution layer, "SA" denotes a self-attention layer, "CA" denotes a cross-attention layer, "DS" and "US" denote down-sampling and up-sampling layers with scale factors equal to 2.}
    \resizebox{\columnwidth}{!}{
    \begin{tabular}{|c|c|c|}
        \hline
         & Number of Feature Channels & Operation  \\
        \hline
         Input & 24 & N/A \\
         \hline
         Block 1 & (32,32,32) & (C, C, DS) \\ 
         Block 2 & (48,48) & (C, DS) \\ 
         Block 3 & (64,64) & (C, DS) \\ 
        \hline
         Block 4 & (64,64,64,64) & (C, SA, CA, DS) \\ 
         Block 5 & (128,128,128,128) & (C, SA, CA, DS) \\ 
         Block 6 & (256,256,256,256) & (C, SA, CA, DS) \\ 
         Block 7 & (256,256,256) & (C, SA, CA) \\ 
         Block 8 & (256,256,256,256) & (C, SA, CA, C) \\ 
         Block 9 & (256,256,256) & (C, SA, CA) \\ 
         Block 10 & (256,256,256,256) & (C, SA, CA, US) \\ 
         Block 11 & (256,256,256) & (C, SA, CA) \\ 
         Block 12 & (256,256,256, 256) & (C, SA, CA, US) \\ 
         Block 13 & (128,128,128) & (C, SA, CA) \\ 
         Block 14 & (128,128,128,128) & (C, SA, CA, US) \\ 
         Block 15 & (64,64,64) & (C, SA, CA) \\ 
         \hline
         Block 16 & (64,64) & (US, C) \\ 
         Block 17 & (48,48) & (US, C) \\ 
         Block 18 & (32,32,14) & (US, C, C) \\ 
         \hline
         Output & 14 & N/A \\
         \hline
    \end{tabular}
    }

    \label{tab:s-na}
\end{table}

\begin{figure}[t!]
    \includegraphics[width=\linewidth, trim=13.25cm 3.8cm 14.25cm 0cm, clip]{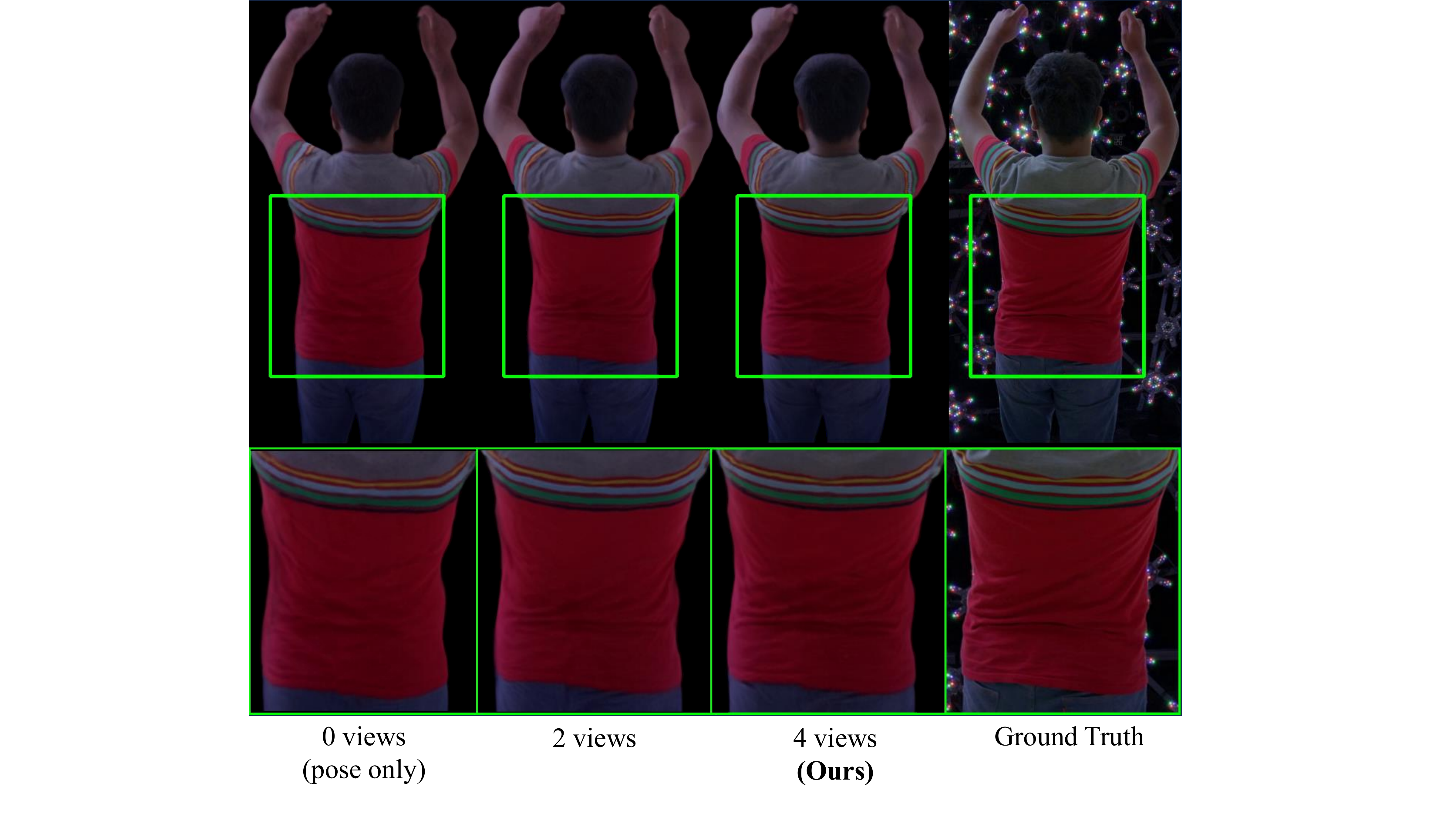}
    \vspace{-18pt}
     \caption{
     \textbf{Effect of the number of input views for our model.}
     As the number of input views decrease, we note the model begins to hallucinate details.
    \vspace{-12pt}
     }
    \label{fig:views}
\end{figure}

\begin{figure}[t!]
    \includegraphics[width=\linewidth, trim=4.25cm 0cm 6.25cm 0cm, clip]{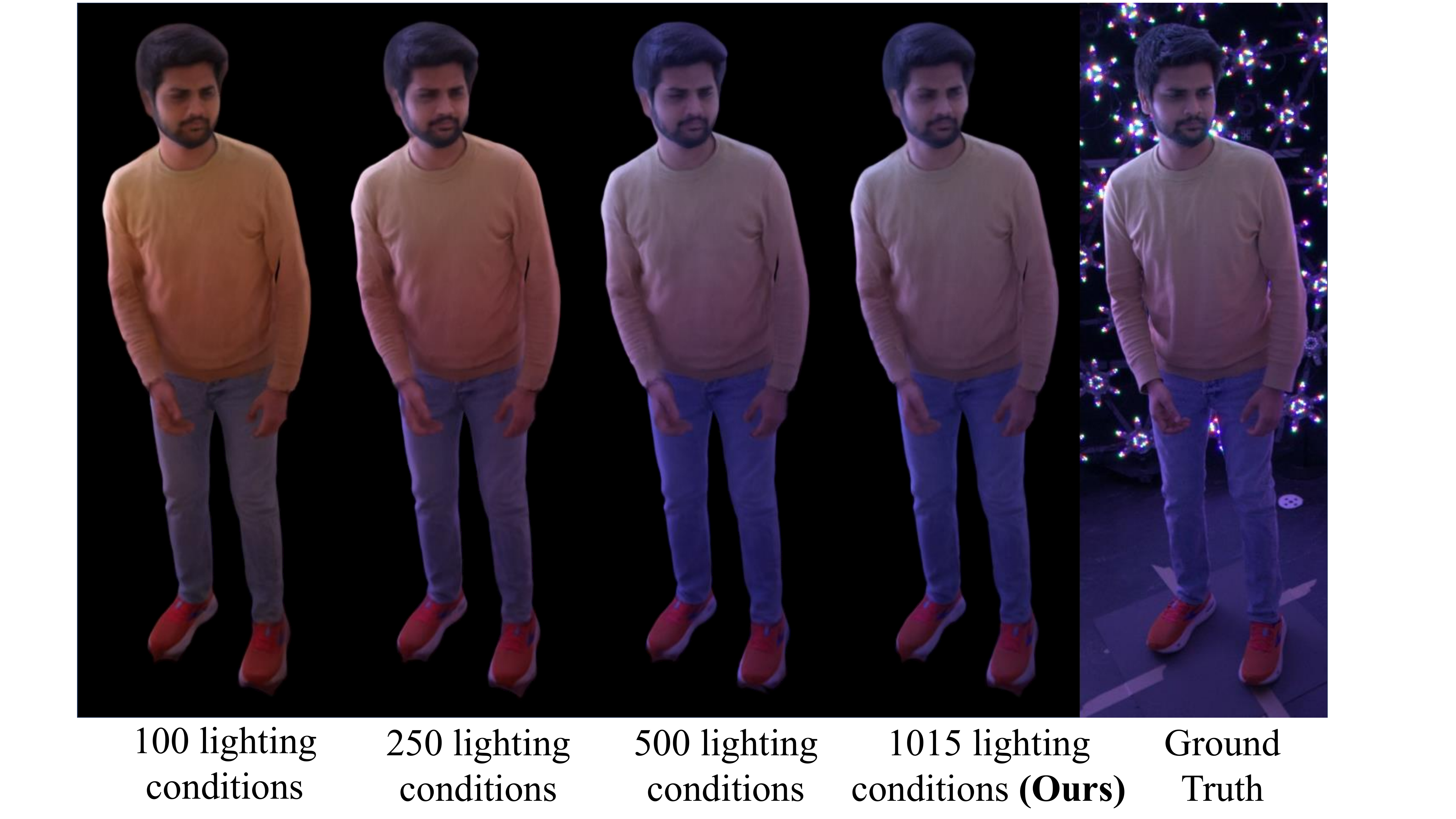}
    \vspace{-18pt}
     \caption{
     \textbf{Effect of the number of lighting conditions our model is exposed to at training time.}
     As the number of conditions decreases, the model overfits and performance drops, highlighting the importance of diverse lighting for end-to-end relighting.
     }
    \label{fig:light}
\end{figure}

\begin{figure}[t!]
    \includegraphics[width=\linewidth, trim=7.25cm 2.8cm 14.25cm 6cm, clip]{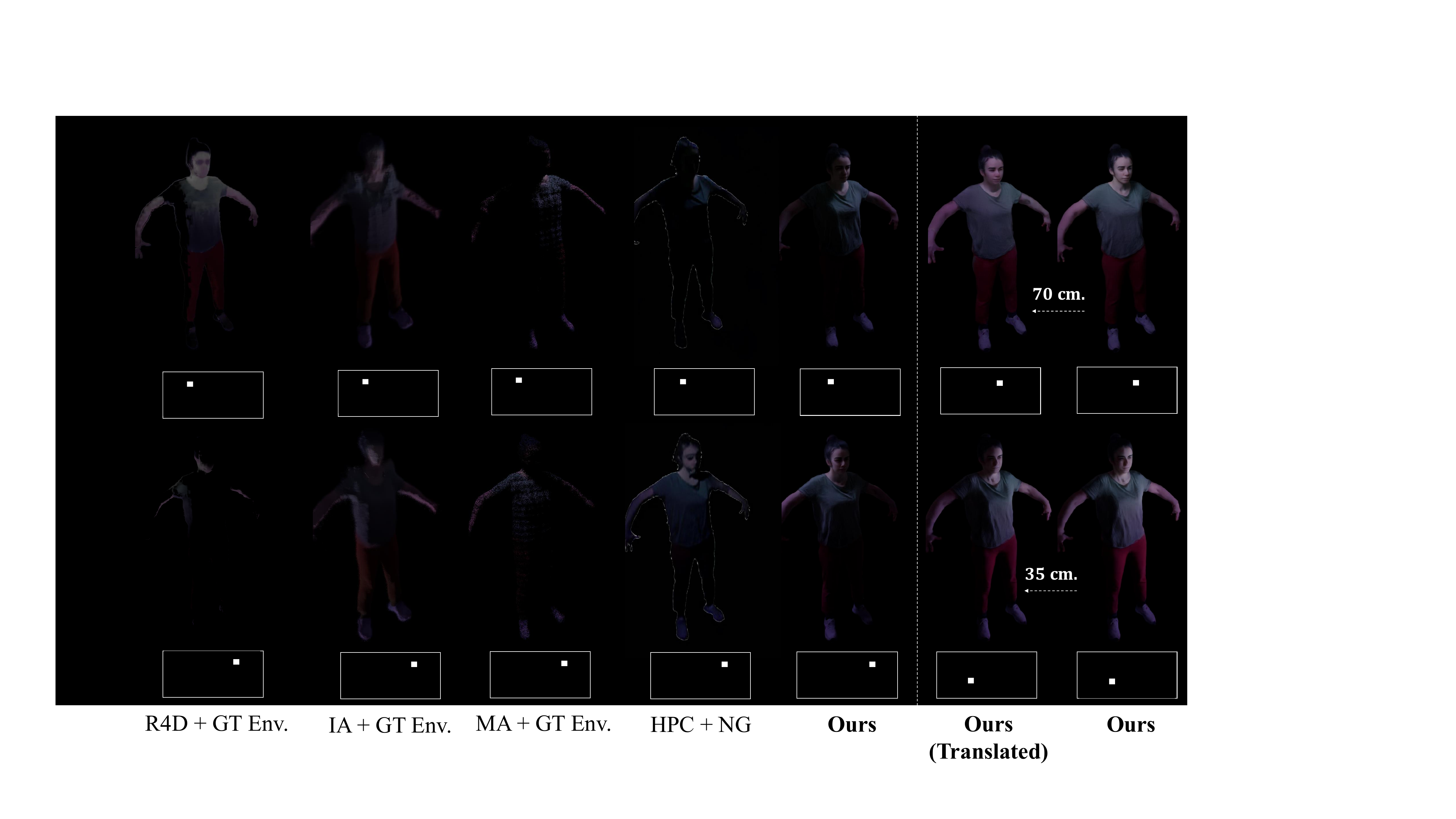}
    \caption{\textbf{OOD comparison.} Here, we compare our method on out-of-distribution lighting conditions, i.e. OLAT environment maps. Notably the model never saw OLAT environment maps during training. Nonetheless, it can generate plausible results while competing methods either produce blurry renderings or completely fail. Moreover, we illustrate that our method can reproduce near field lighting effects by translating the human by 35cm, i.e. modifying the positional map and diffuse shading, and we can observe a plausible change in illumination.}

     \vspace{-10pt}
    \label{fig:exp-olat_near}
\end{figure}

\section{Training Details}\label{sec:s-loss}

\textbf{\textit{Character Animation Networks} $\mathcal{G}$} consists of two graph neural networks $\mathcal{G}_\mathrm{eg}, \mathcal{G}_\mathrm{delta}$ trained separately in uniformly lit frames.
In particular, $\mathcal{G}_\mathrm{eg}$ is supervised with silhouette loss over multiview segmented images, and chamfer loss over NeuS2~\cite{wang2023neus2} reconstructed point clouds.
Since embedded graph deformation has a lower degree of freedom, we introduce an as-rigid-as-possible loss as a regularization.
\begin{equation}
    L_\mathrm{EG} = L_\mathrm{Sil} + L_\mathrm{Chamfer} + L_\mathrm{ARAP}
\end{equation}
The per-vertex offset is driven by additional photometric losses, with stronger regularization losses including Laplacian loss and Isometry loss.
\begin{equation}
    L_\mathrm{Delta} = L_\mathrm{Sil} + L_\mathrm{Photo} + L_\mathrm{Chamfer} + L_\mathrm{Lap} + L_\mathrm{Iso}
\end{equation}
\textbf{\textit{AlbedoNet} $\mathcal{H}$} is trained with dense-view image capture in uniformly lit frames using photometric losses.
Consequently, similar to the high-frequency normal estimation $\hat{\boldsymbol{n}}$, the albedo feature $\hat{\boldsymbol{\rho}}$ incurs a one-frame delay when incorporated into the training process of our neural relightable model for subsequent relit frames.
\textit{AlbedoNet} is supervised with $L_1$ loss over multiview image captures in the uniformly lit frames.
Unlike Holoported Character~\cite{shetty2024holoported}, we leverage a one-stage UNet architecture without additional super-resolution.
The multiview images are resized into $540\times1024$ as it is sufficient to represent high-quality albedo features in a $512\times512$ texture map.

\noindent \textbf{\textit{RelightNet} $\mathcal{F}$} is supervised with multi-view images. 
Unlike the Character Animation Networks $\mathcal{G}$ and AlbedoNet $\mathcal{H}$ that are trained with uniformly lit frames, \textit{RelightNet} is trained with the interleaved relit frames using photometric losses and regularization terms reweighted by $\lambda$: $L_\text{train} = \lambda_1L_\text{1} + \lambda_2L_\text{reg} + \lambda_3L_\text{vgg}$.
Here, $L_\text{1}$ is the mean of the pixel-wise $L_\text{1}$ distance between the rasterized image $\boldsymbol{I}_\mathrm{pred}$ and the ground-truth image $\boldsymbol{I}_\mathrm{gt}$. 
$L_\text{vgg}$ is the VGG perceptual loss~\cite{johnson2016perceptual}.
The regularization term $L_\text{reg}$ penalizes the deviations of scaling offsets $\mathbf{s}_i$ from 1 and position offsets $\delta\mathbf{p}_i$ from 0. 
However, the training of $\textit{RelightNet}$ can diverge in the early stage due to the undesired prediction of opacity and scaling parameters.
To stabilize training, we further introduce a warm-up phase where we regularize the initial Gaussian prediction with Eq.~\ref{eq:objective_warmup}.

\begin{align}
L_{\text{Warmup}} = \frac{1}{N_G} \sum_{i=1}^{N_G} \big( 
    &\lambda_\text{s}\|\boldsymbol{s}_i - 1\|_2^2
    + \lambda_\text{t}\|\delta \boldsymbol{p}_i\|_2^2 \notag \\
    &+ \lambda_\text{o}\|o_i - o_0\|_2^2 
    + \lambda_\text{c}\|\boldsymbol{c}_i - \boldsymbol{c}_i^\text{tem}\|_2^2 \big),
\label{eq:objective_warmup}
\end{align}
where $\boldsymbol{c}_i^\text{tem}$ is the color of the texture map of the template mesh captured in a fully lit environment, and $o_0=0.95$ is the default opacity. 
In this stage, the input environment map $\boldsymbol{E}$ is set as a uniformly lit environment map.
This loss is linearly switched to $L_\text{train}$ between $25,000$ and $26,000$ iterations.

We train RelightNet for $360,000$ iterations on four H100s with a batch size of 4 and accumulate gradients every second iteration. 
We use the ADAM optimizer with a learning rate set to a constant 1e-4. 
For hyperparameters, we use $\lambda_1 = 2.7$, $\lambda_2 = 1.2$, $\lambda_3 = 75$, $\lambda_\text{scale} = 1$, $\lambda_\text{pos} = 1$, $\lambda_\text{s} = 1$, $\lambda_\text{t} = 0.1$, $\lambda_\text{o} = 1$, $\lambda_\text{c} = 1$.

\section{Additional Data Capture Details} \label{sec:s-data}
The lightstage we use is equipped with 331 calibrated LED panels and 40 cameras, operating at 60 \textit{fps} with 4K resolution.
In total, we capture five sequences, each featuring a different identity and unique clothing. 
Each subject was captured twice: once for training and once for testing, ensuring that the test set contains novel poses. %
We sampled 1,015 environment maps from the Laval Indoor HDR dataset \cite{gardner2017learning} for the training set, and held 8 environment maps out for the test set.
These environment maps were downsampled to a resolution of $256 \times 512$, with each pixel mapped to the nearest LED panel in 3D space.
The intensities of pixels mapped to the same LED panel were averaged, and the resulting environment map was linearly normalized to match the maximum intensity that the LED panels could simulate. 
These processed environment maps are considered the ground-truth environment maps for all experiments.
\section{Baseline Implementation Details}\label{sec:s-baseline}
\textbf{Relighting4D.}
\begin{figure}[t!]
    \includegraphics[width=\linewidth, trim=9cm 0cm 4.25cm 0cm, clip]{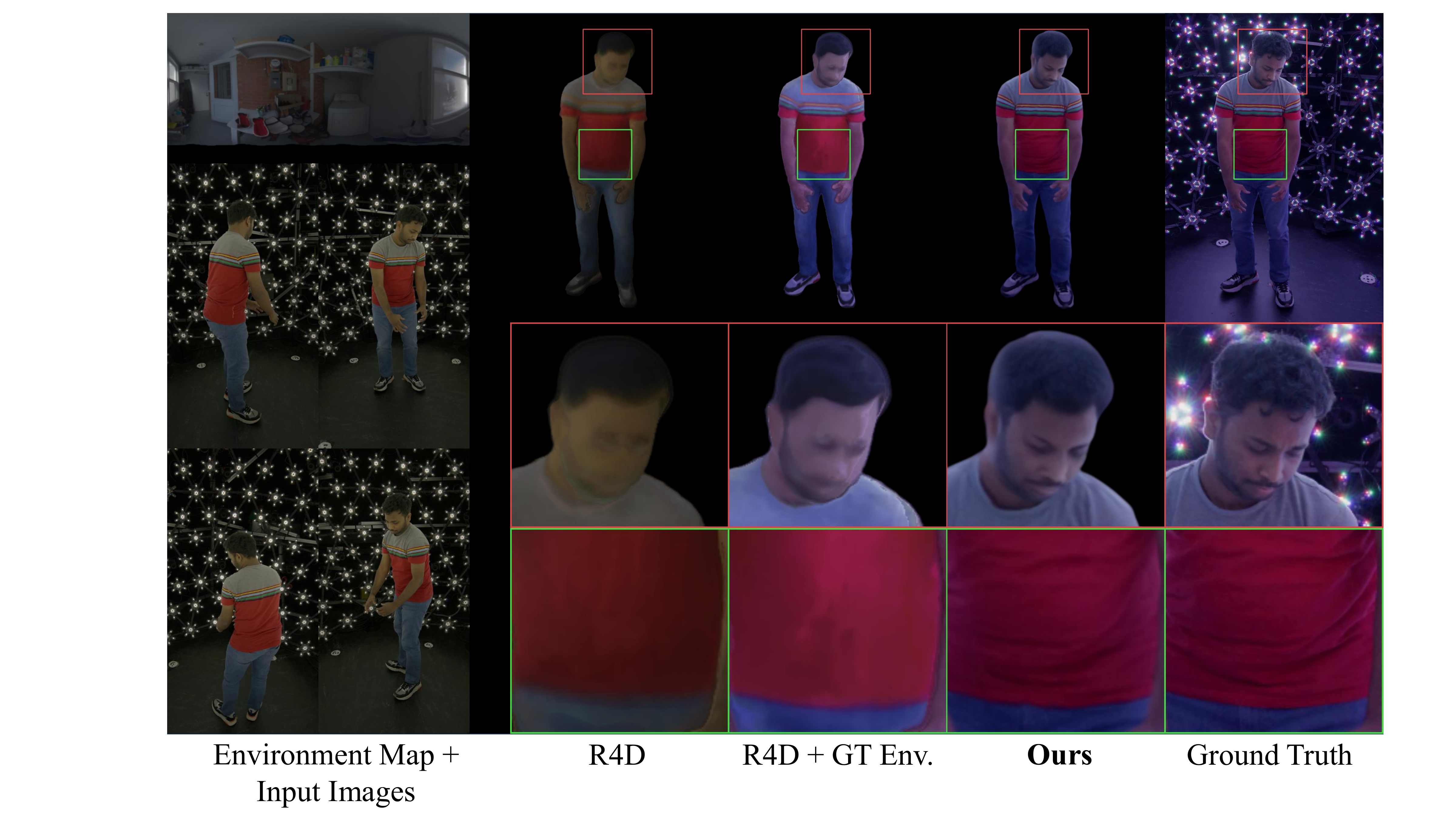}
    \vspace{-18pt}
     \caption{
     \textbf{Effect of using diverse illumination and ground truth environments for training.}
     On the left, we show test results of R4D~\cite{chen2022relighting4d} trained under uniform illumination only. Next, we show the same baseline trained with diverse illumination conditions using the ground truth environment maps (R4D + GT Env). We note significant generalization improvement for R4D with this training strategy.
    \vspace{-12pt}
     }
    \label{fig:r4d}
\end{figure}

Relighting4D~\cite{chen2022relighting4d} is originally designed for \textit{replay} of human performance captured from a single-view camera, which optimizes a time-dependent latent vector that spans the training sequence.
We extend Relighting4D as an animatable and relightable approach trained with multiview supervision. 
To enable the generalization ability to novel motion sequences, in each test frame, we provide the closest latent vector from the training sequence. 
In addition, we replace the original used SMPL~\cite{loper2023smpl} geometry with our high-quality mesh-based character model for a fair comparison.
The Relighting4D is trained with uniformly lit frames, which contain a static illumination with a wide variety of motions.
At test time, we scale our normalized environment maps using an optimized radiance scaling factor. 

\textbf{Relighting4D + GT Env Map.}
For this variant of Relighting4D, we use uniformly lit frames to train the geometry features of Relighting4D following the same training scheme as before.
For the material features estimation, we utilize the supervision of the relit frames.
Instead of jointly optimizing the full environment map with material parameters, this baseline takes the ground truth environment map as input, while optimizing one scaling factor for each scene, which rescales the normalized environment map to be compatible with the underlying BRDF model.

\textbf{IA + GT Env Map.}
We train Intrinsic Avatar in a similar way to R4D + GT Env. Note that, unlike Relighting4D, it is non-trivial for us to upgrade the underlying mesh with our tracking result in IA, as IA is designed with strong entanglement with SMPL articulation. Therefore, we rely on the original method's SMPL tracking. We also note that due to a bug in the code, the pose correction module was not working and had to be disabled during training.

\textbf{MA + GT Env Map.}
We train Mesh Avatar similarly to IA + GT Env. For this baseline, we train directly with the relit frames, since it performs geometry optimization along with material parameter optimization. 

\textbf{Holoported Characters.}
Holoported Characters~\cite{shetty2024holoported} produces human rendering under similar lighting conditions at training and test times. 
It is trained using the multiview images from uniformly lit frames for supervision. 
In the test time, we leverage geometry features and mesh from the relit frame and the partial texture of the prior uniformly lit frame, strictly following our testing setup.

\textbf{Holoported Character + Neural Gaffer.}
Neural Gaffer~\cite{jin2024neural} is a generative image-to-image relighting method. 
It is a pretrained model that takes an input image and an environment map to provide a relit image. 
We finetune Neural Gaffer for each identity separately and use it to translate the novel view rendering of Holoported Characters to obtain a relit output. 
Since the environment map in Neural Gaffer is considered object-centric, for each test camera, we transform the global space environment into camera space accordingly.

\begin{figure}[t!]
    \includegraphics[width=\linewidth, trim=54.75cm 15cm 0cm 0cm, clip]{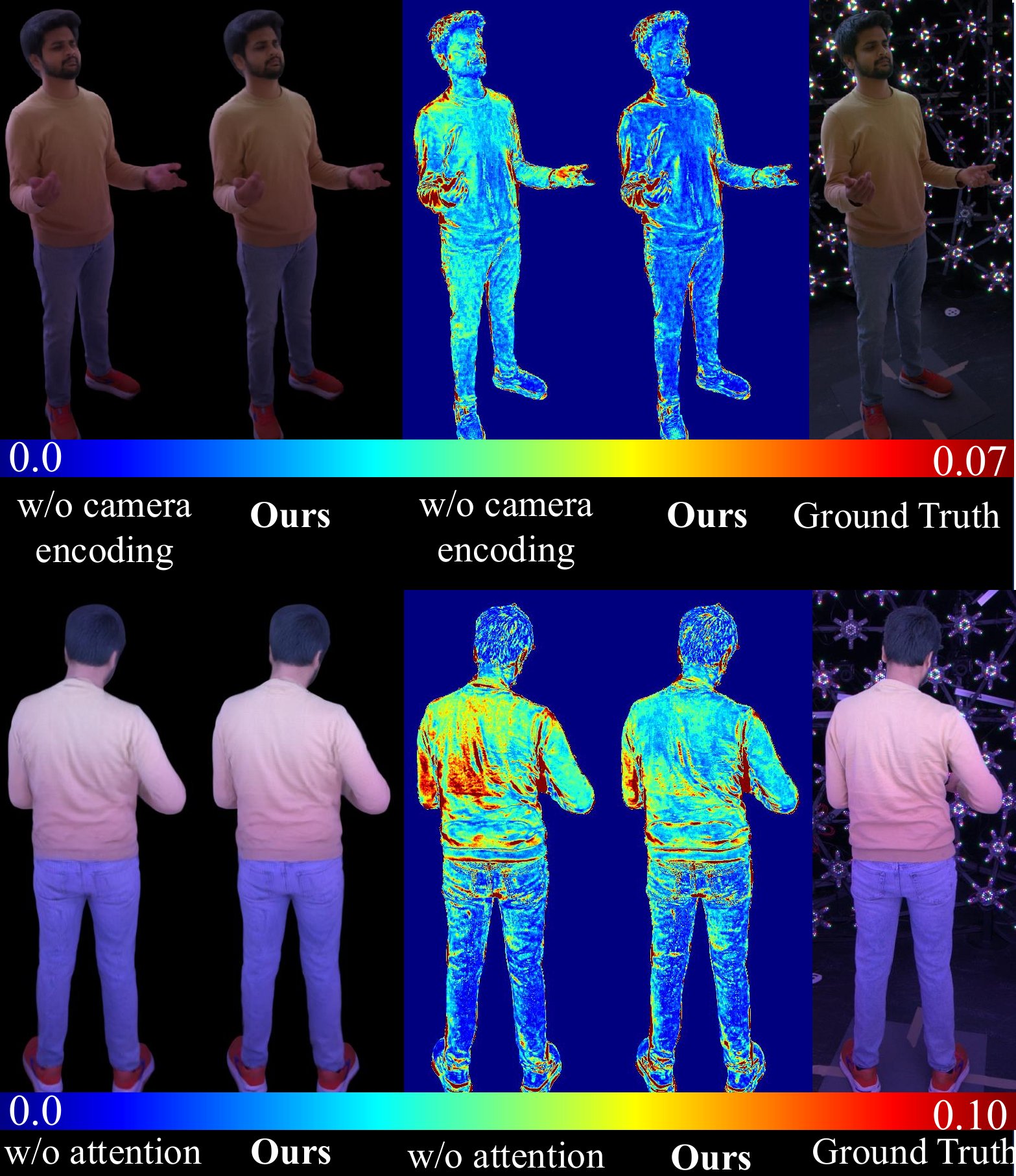}
     \caption{
     \textbf{Additional qualitative ablations.} Following Fig. 6 in the main paper, we demonstrate additional visual ablation results. By removing the camera encoding, our model is no longer able to learn view-dependent effects. Without the attention mechanism, the model is unable to learn the correct correlation with the environment. 
     }
    \label{fig:more-ablation-results}
\end{figure}

\begin{figure}[t!]
    \includegraphics[width=\linewidth, trim=0cm 0cm 54.5cm 19cm, clip]{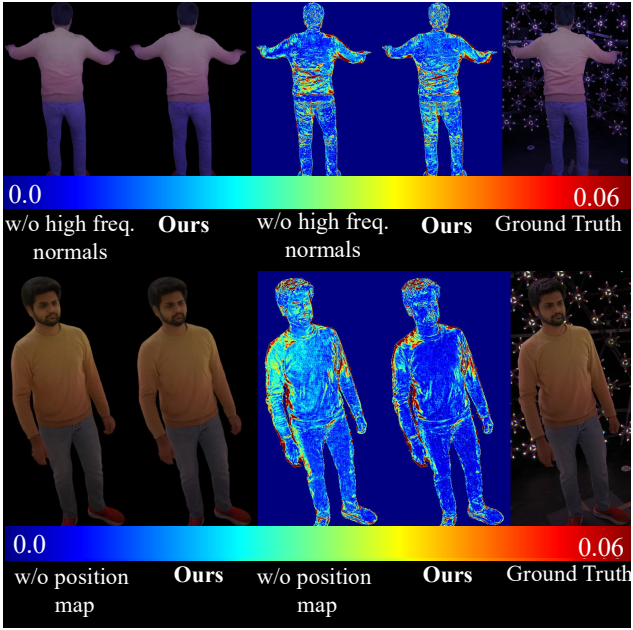}
     \caption{
     \textbf{Additional qualitative ablations.} Following Fig. 6 in the main paper, we demonstrate additional visual ablation results. Without the high-frequency normals, our method is unable to model wrinkle details correctly. Excluding the position map leads to incorrect relighting, especially when the subject approaches a light source. 
     }
    \label{fig:more-ablation-results2}
\end{figure}

\section{Dynamic OLAT Experiment}\label{sec:s-olat}
Since OLAT images are darker than environment-lit images, we add structured supervision to stabilize training, using coarse UV-space Microfacet rendering.
We first optimize albedo and roughness maps on the DDC-tracked mesh using the Microfacet BRDF formulation of~\cite{chen2022relighting4d}, supervised with ground-truth relit data.  
We compute per-texel normals, per-light visibility, and optimize a scaling parameter for the environment map.  
The resulting maps are used to render a coarse UV-space image, which serves as the input to the network in place of the diffuse shading feature.
Instead of directly predicting relit colors, the network learns to predict additive color deltas over the coarse rendering.
Evaluation follows our standard protocol with three novel views, novel poses, and eight held-out OLAT lighting conditions.
At test time, we render the full OLAT basis and linearly combine them to reconstruct target environment maps.

Since the OLAT frames were extremely dark, we had to use a frame interpolation network~\cite{zhang2024vfimamba} to generate intermediate uniformly lit frames on which we compute the foreground masks. Also, compared to 360,000 iterations of our model, this model took 1,200,000 iterations to converge. 

\section{Runtime and Memory}
\label{sec:s-runtime}
Tab.~\ref{tab:runtime} reports a per-component runtime breakdown (H100; averaged over 100 iters; no data loading). The main bottleneck is Sapiens normal estimation (amenable to distillation or lighter model). Tab.~\ref{tab:baseline_runtime} compares runtime and memory footprint against baselines. RelightNet has $\approx$27M parameters.
Overall, our full pipeline runs at $\approx$2\,FPS, substantially faster than radiance-field--based R4D/IA (seconds per frame) and HPC+NG (multiple diffusion denoising steps), but slower than MA (7\,FPS); a lightweight variant of our model without the Sapiens normal prior reaches up to 9\,FPS.

\begin{table}[h]
\footnotesize
\centering
\vspace{-6pt}

\centering
\captionof{table}{\textbf{Runtime Breakdown}}
\label{tab:runtime}
\vspace{-6pt}
\setlength{\tabcolsep}{3pt}
\renewcommand{\arraystretch}{1.05}
\begin{tabular}{|c|c|}
\hline
Component & Runtime (ms.) \\
\hline
Character Animation Module & 9.41 \\
Sapiens & 244.10 \\
AlbedoNet & 15.31 \\
Diffuse Shading (Open3D/Optix) & 163.60 / 26.73 \\
Texture Unprojections & 46.78 \\
RelightNet + Rasterization & 30.50 \\
\hline
\textbf{Total (Open3D/Optix)} & \textbf{509.70 / 372.83} \\
\hline
\end{tabular}

\captionof{table}{\textbf{Runtime/Memory vs. Baselines}}
\label{tab:baseline_runtime}
\begin{tabular}{|c|c|c|}
\hline
Method & sec. & MB \\
\hline
R4D & 8.49 & 8399 \\
IA & 32.72 & 9181 \\
MA & 0.15 & 23407 \\
HPC & 0.08 & 40784 \\
HPC+NG & 2.13 & 45831 \\
Ours & 0.51 & 49884 \\
\hline
\end{tabular}
\end{table}

\vspace{-8pt}
\section{Further Ablations}
\label{sec:s-fur_abl}
\begin{figure}[t]
    \includegraphics[trim = 27cm 10cm 8cm 6cm , clip , width=\linewidth]{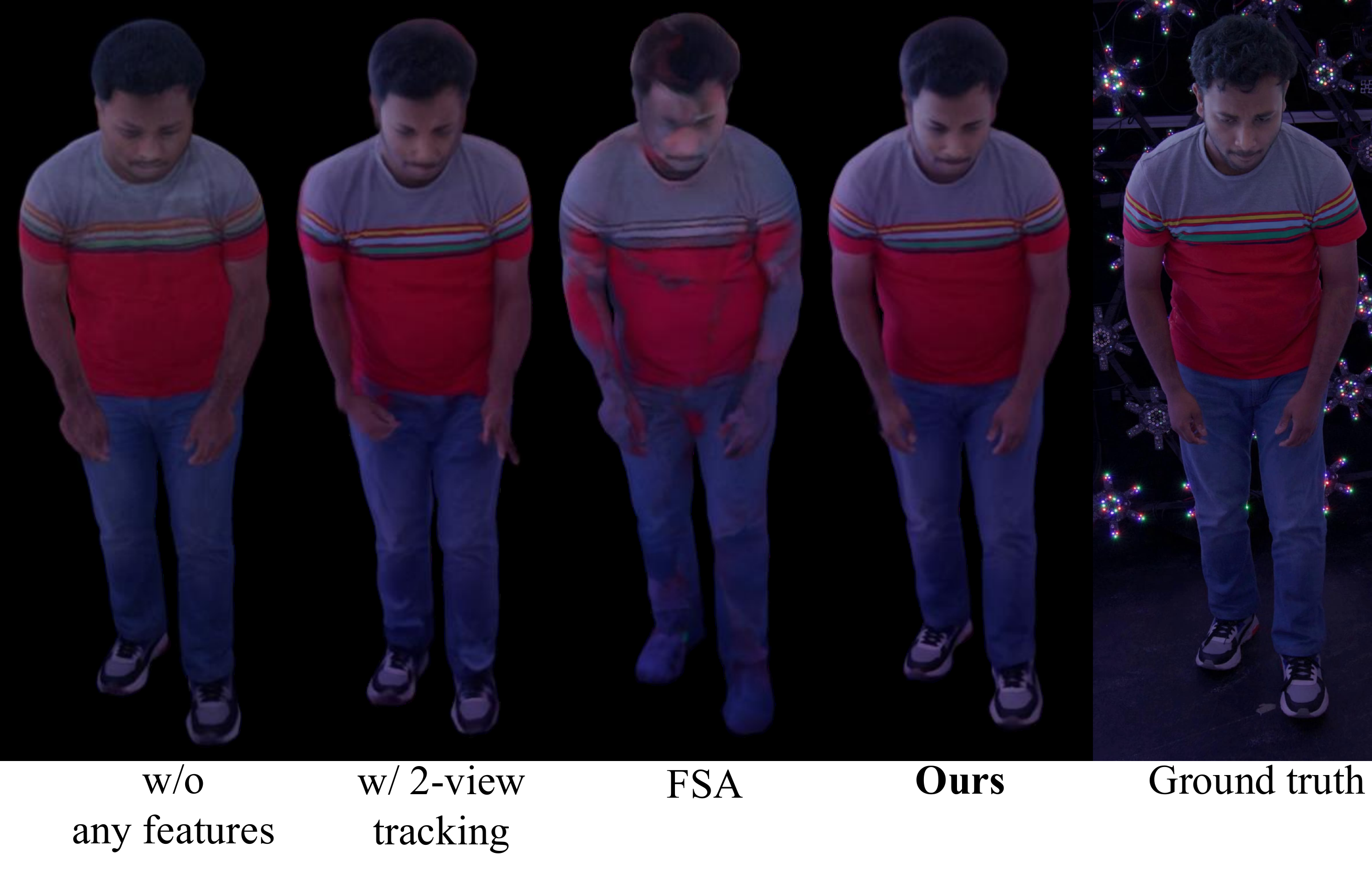}
    \vspace{-20pt}
     \caption{
    \textbf{Additional qualitative ablations.} We provide further visual ablation results. Removing all physics-informed features causes a significant drop in rendering quality. Limiting tracking and image observations to only two views primarily reduces performance due to inaccurate tracking. Finally, few-shot adaptation of RelightNet introduces artifacts arising from the material and UV bias of the original subject.
    \vspace{-12pt}
     }
    \label{fig:abl_all}
\end{figure}

We perform additional ablations and present both quantitative (Tab.~\ref{tab:abl_all}) and qualitative results (Fig.~\ref{fig:abl_all}).

\textbf{Importance of Physics-informed Features.}
\label{s-imp_phy}
We ablate all physics-informed features. Performance drops significantly when removing them (Tab.~\ref{tab:abl_all}, Fig.~\ref{fig:abl_all}), confirming that our gains arise from the combination of physics-informed features, network design, and capture strategy.

\textbf{Sparse-View Tracking.}
\label{s-more_sparse}
Using skeleton tracking from four sparse views slightly degrades performance but still yields high-quality renderings (Supplementary Video); tracking error w.r.t. dense-view ground truth is MPJPE $=18.54\,$mm, mostly from hands. With two-view tracking integrated with our 2-view ablation (Tab. 2 of main paper), tracking errors increase substantially, but we still observe visually coherent renderings in many cases (Tab.~\ref{tab:abl_all}, Fig.~\ref{fig:abl_all}), suggesting feasibility of a setup with reduced input requirements, with improved tracking or learned pose priors.

\textbf{Few-shot Adaptation (FSA).}
\label{s-fsa}
We finetune a Subject-2 checkpoint on 12 Subject-5 samples (diverse poses/lighting/viewpoint), updating Input/Output blocks and Blocks 1–3, 16–18 (Tab.~\ref{tab:s-na}) for 1k iterations. The adapted model shows artifacts (Tab.~\ref{tab:abl_all}, Fig.~\ref{fig:abl_all}) due to Subject-2 material/UV bias. We acknowledge this as a main limitation and consider it for future work, since recent developments in generalizable priors show great promise.

\begin{table}[t]
\footnotesize
\centering
\caption{\textbf{Further Ablations.} We ablate the need for physics-informed features, evaluate performance when both, skeleton tracking and image inputs are limited to just two views, and assess RelightNet’s cross-identity generalization by measuring few-shot adaptation performance.}
\label{tab:abl_all}
\vspace{-6pt}
\begin{tabular}{|c|c|c|c|}
\hline
Method & PSNR $\uparrow$ & LPIPS $\downarrow$ & SSIM $\uparrow$ \\
\hline
w/o any features & 29.68 & 7.65 & 84.45 \\
w/ 2-view tracking & 27.82 & 10.25 & 81.35 \\
w/ few-shot adaptation (FSA) & 29.07 & 10.09 & 83.72 \\
\hline
\textbf{Ours} & \textbf{32.07} & \textbf{5.55} & \textbf{89.34} \\
\hline
\end{tabular}
\end{table}

\section{Additional Qualitative Results}\label{sec:s-qual}
\begin{figure*}[h!]
    \centering \includegraphics[width=0.9\linewidth, trim = 17cm 0cm 30cm 0cm, clip]{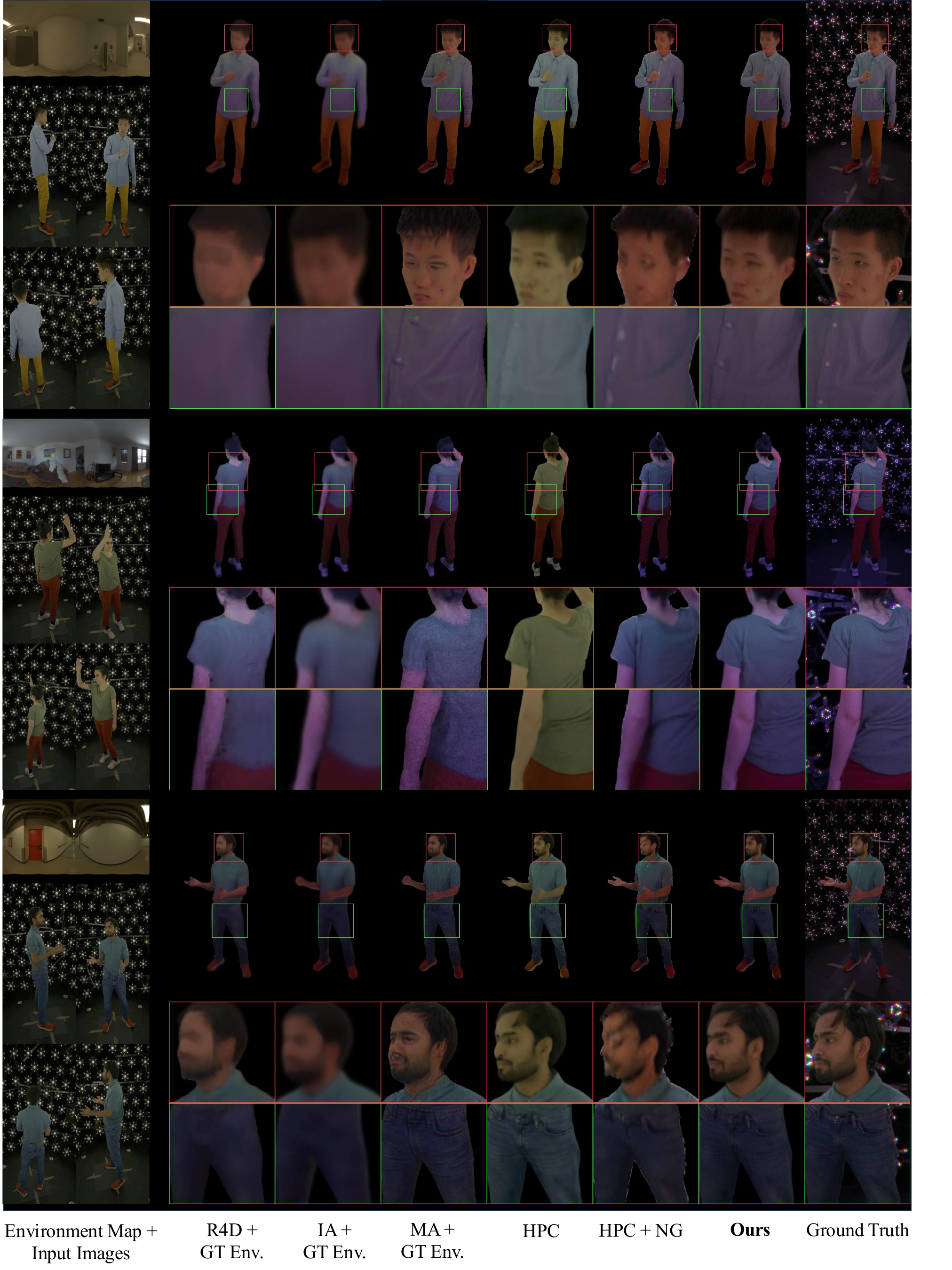}
    \caption{\textbf{Additional Qualitative Comparisons.} We show additional visual comparison between our proposed baseline methods, \textit{i.e.} 
    R4D~\cite{chen2022relighting4d}, IA~\cite{wang2024intrinsicavatar} and,
    MA~\cite{chen2024meshavatar} and,
    HPC~\cite{shetty2024holoported} with/without NG~\cite{jin2024neural} postprocessing on three other subjects. This figure showcases the significant improvement of our method in rendering and relighting quality among all subjects, which better supports our conclusion drawn in Fig. 5 of the main paper.}
    \vspace{-10pt}
    \label{fig:s-more-results}
\end{figure*}

\begin{figure*}[h!]
    \centering \includegraphics[width=.9\linewidth, trim=13cm 0cm 32cm 0cm, clip]{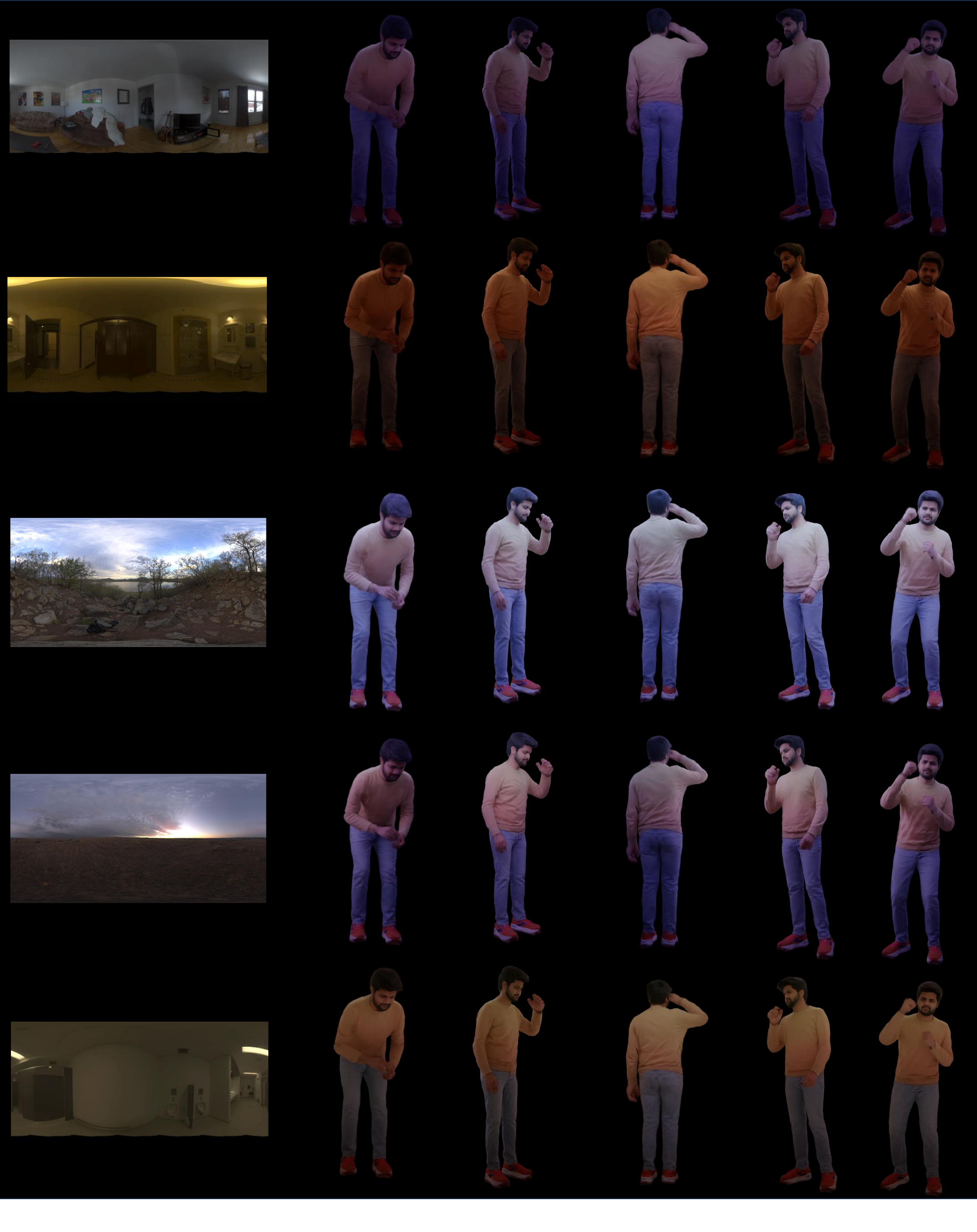}
    \caption{\textbf{Qualitative Results in Novel Illumination.} We present additional qualitative results under versatile unseen environments, including two outdoor environments downloaded online. Our model can generate realistic shading, including self-shadows, while preserving visual details under novel illuminations with different brightness, color tone, etc.}
    \label{fig:novel-light-results}
\end{figure*}

We conduct extensive experiments with qualitative results to demonstrate the superiority and robustness of our approach.

\textbf{Relighting4D Training Strategy.}
In Fig.~\ref{fig:r4d}, we show the effect of training the Relighting4D~\cite{chen2022relighting4d} baseline with uniformly lit frames vs re-lit frames with ground truth environment maps. We observe more generalisable optimisation of material parameters by training with the re-lit frames and ground truth environment maps input.

\textbf{Impact of Lighting Diversity.}
\label{sec:div}
We analyze the impact of our data capture strategy for dynamic human avatar relighting.
To assess lighting diversity, we progressively reduce the number of environment maps a subject is exposed to (Tab.~\ref{tab:light}, Fig.~\ref{fig:light}).  
Relighting quality clearly degrades with fewer lighting conditions, showing that diverse illumination is crucial for learning accurate relighting.

\textbf{Number of Views.}
In Fig.~\ref{fig:views}, we show the effect of the number of views our model uses on the final rendering quality. As decreasing the number of views leads to more information being occluded, we observe lower reconstruction fidelity of high-frequency details as well as more hallucination.

\textbf{Generalization to OLAT and Near Field Relighting.}
We test the robustness of our method in challenging lighting conditions, \textit{i.e.} OLAT and near-field lighting, which is out of the distribution of our training environments (see Fig.~\ref{fig:exp-olat_near}).
Without explicit modeling of light linearity, our learning-based \textit{RelightNet} can still generate plausible shading for these illuminations. 
In contrast, NG fails to disentangle identity and shading in the OLAT setting, R4D + GT Env. based on empirical BRDF models can produce reasonable but imprecise OLAT effects. IA/MA + GT Env. lead to artifacts similar to R4D + GT Env.
This demonstrates that our models learns the correct light-material disentanglement, even without being trained on OLAT data.

\textbf{Ablations.}
\begin{table}[t]
    \footnotesize 
    \centering
    \caption{
    \textbf{Impact of Lighting Diversity.}
    We ablate different choices of our data capture strategy. 
    }
    \label{tab:light}
    \vspace{-5pt}
    
    \setlength{\tabcolsep}{1.2pt} 
    \renewcommand{\arraystretch}{1.2} 
    \begin{tabular}{|c?c|c|c|}
         \hline
         Number of lighting conditions & PSNR $\uparrow$ & LPIPS  $\downarrow$ &SSIM $\uparrow$\\
         \hline
         100 & 28.79 & 9.12 & 88.70 \\
         250 & 29.91 & 7.71 & 88.95 \\
         500 & \uline{30.91} & \textbf{6.97} & \uline{89.54} \\
     
         \hline
         \textbf{1015 (Ours)} & \textbf{31.38} & \uline{7.01} & \textbf{90.00} \\
         \hline

    \end{tabular}
    \vspace{-10pt}

\end{table}

In Fig.~\ref{fig:more-ablation-results} and Fig.~\ref{fig:more-ablation-results2}, we provide extra ablation results following Fig. 6 of the main paper to validate other design choices.
We are able to see the visual impact of our design decisions, demonstrating that each component plays an important role in learning the correct light transport.

\textbf{Miscellaneous Results.}
In Fig.~\ref{fig:s-more-results}, we show additional qualitative comparisons, where our method constantly achieves superior results with sharper clothing and facial details, with more accurate color shading. 
In Fig.~\ref{fig:novel-light-results}, we test the robustness of our relightable avatar against novel illuminations in addition to OLAT/near-field lightings, including outdoor environments downloaded from Poly Haven\footnote{https://polyhaven.com/}.
Our proposed method renders consistent and sharp visual details with realistic shadings corresponding to input environment maps.

\end{document}